\PassOptionsToPackage{table}{xcolor}
\documentclass[10pt,twocolumn,letterpaper]{article}

\usepackage{times}
\usepackage{epsfig}
\usepackage{graphicx}
\usepackage{amsmath}
\usepackage{amssymb}
\usepackage[inkscapeformat=png]{svg}
\usepackage{multirow}
\usepackage[title]{appendix}
\usepackage{graphicx}
\usepackage{subcaption}
\usepackage{romannum}
\usepackage{mathtools}

\newcommand{\normalx}{\ensuremath{\,\textrm{x}}}

\usepackage[breaklinks=true,bookmarks=false]{hyperref}

\begin{document}

\title{Pre-NeRF 360: Enriching Unbounded Appearances for Neural Radiance Fields}

\author{Ahmad AlMughrabi \and Umair Haroon \and Ricardo Marques$^*$ \and Petia Radeva$^*$ \and \\
Universitat de Barcelona}
\date{}
\maketitle

\begin{abstract}
   Neural radiance fields (NeRF) appeared recently as a powerful tool to generate realistic views of objects and confined areas. Still, they face serious challenges with open scenes, where the camera has unrestricted movement and content can appear at any distance. In such scenarios, current NeRF-inspired models frequently yield hazy or pixelated outputs,
   suffer slow training times, and might display irregularities, because of the challenging task of reconstructing an extensive scene from a limited number of images. We propose a new framework to boost the performance of NeRF-based architectures yielding significantly superior outcomes compared to the prior work. Our solution overcomes several obstacles that plagued earlier versions of NeRF, including handling multiple video inputs, selecting keyframes, and extracting poses from real-world frames that are ambiguous and symmetrical. Furthermore, we applied our framework, dubbed as "Pre-NeRF 360", to enable the use of the Nutrition5k dataset in NeRF and introduce an updated version of this dataset, known as the N5k360 dataset. The source code, the dataset, and pre-trained weights for Pre-NeRF are publicly available at\footnote{\href{https://amughrabi.github.io/prenerf}{https://amughrabi.github.io/prenerf}}.
   \def\thefootnote{*}\footnotetext{These authors contributed equally to this work.}
\end{abstract}

\section{Introduction}
Addressing the persistent issue of view synthesis in 3D scene reconstruction and rendering using 2D images has received a lot of attention in the literature \cite{barron2021mip, barron2022mip, mildenhall2022nerf, mildenhall2021nerf, verbin2022ref}. Recently, neural volumetric representations, specifically Neural Radiance Fields (NeRF) \cite{mildenhall2021nerf}, emerged as a promising approach for learning to represent 3D objects and scenes from an input set of 2D images. After the training phase, NeRF is able to map 5D input coordinates (consisting of a position in 3D scene-space and a 2D spherical direction representing the radiance flow direction) to scene properties such as color and density. This radiance field information is then used to synthesize novel views of the learned 3D scene by resorting to direct volume rendering techniques.

Several methods based on NeRF were proposed to improve view synthesis. For instance, these NeRF variations methods differ in many aspects, such as positional encoding \cite{barron2021mip}, sampling strategy \cite{yu2021pixelnerf}, network architecture \cite{barron2022mip}, etc. Yet, these NeRF variations methods aggregate colors and densities of discrete points along viewing rays via differentiable volumetric rendering to synthesize novel views termed NeRF-based methods \cite{mildenhall2021nerf}.

\begin{figure}[t]
\setlength{\tabcolsep}{1pt}
\scriptsize
\captionsetup[subfigure]{justification=centering}
\centering
\begin{center}
\begin{tabular}{cc}
    \begin{subfigure}[b]{.5\linewidth}
         \includegraphics[width=\textwidth]{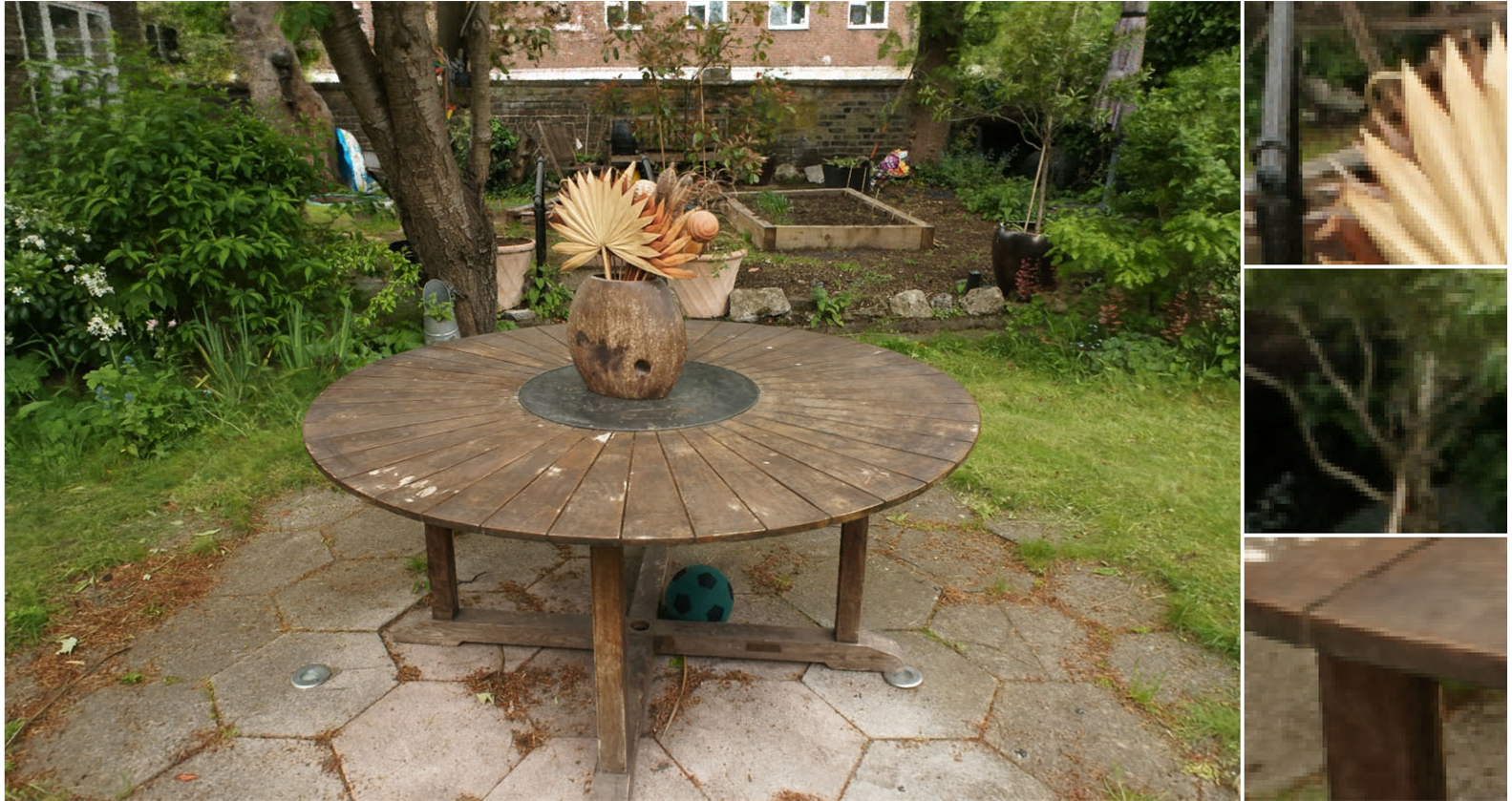}
         \setcounter{subfigure}{0}%
         \raggedleft\caption{Mip-NeRF 360 \cite{barron2022mip}
         }
         \label{fig:mip-nerf-360}
     \end{subfigure} & 
     \begin{subfigure}[b]{.5\linewidth}
         \includegraphics[width=\textwidth]{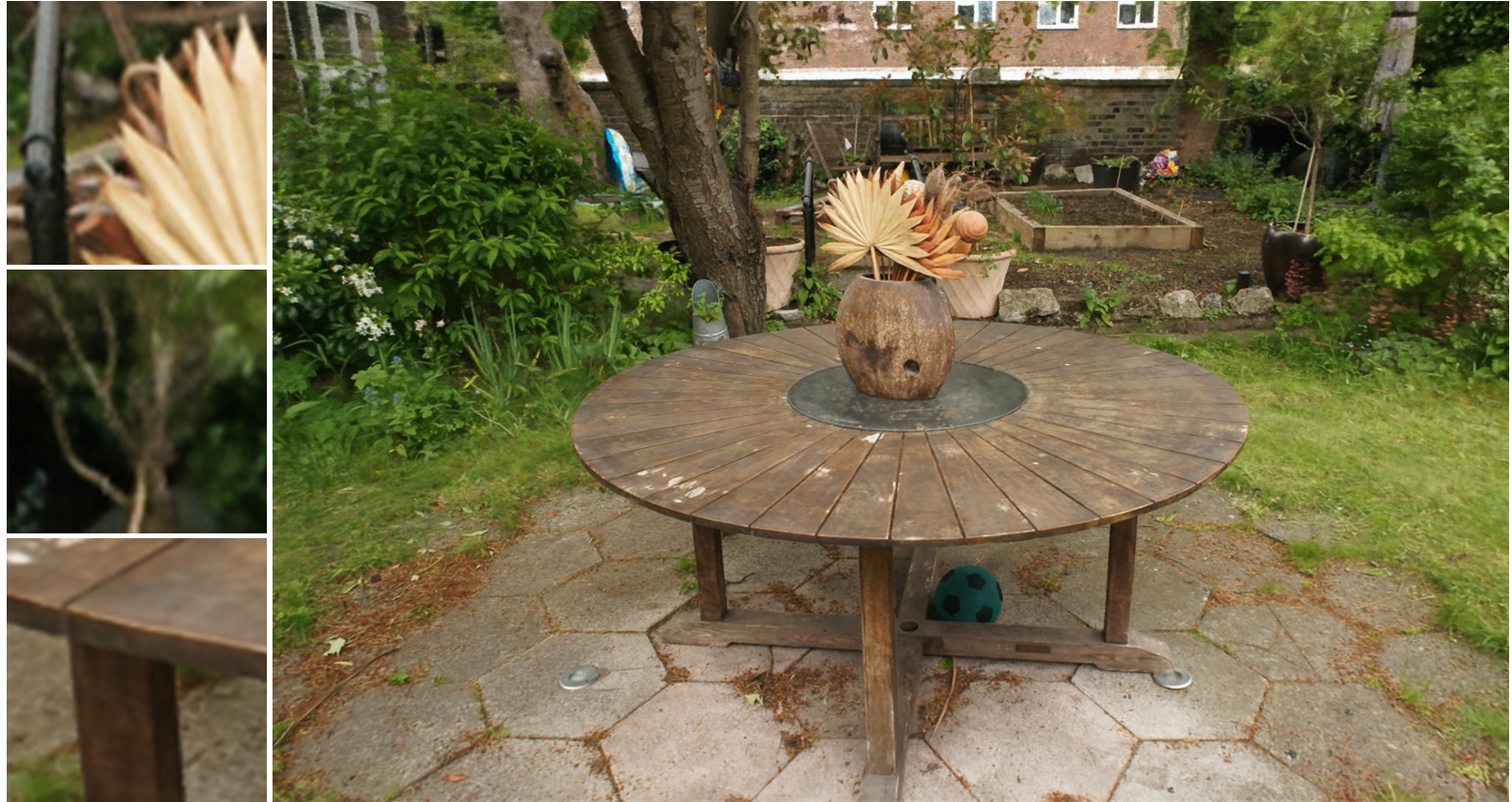}
         \setcounter{subfigure}{1}%
         \raggedleft\caption{Pre-NeRF(Ours)
         }
         \label{fig:pre-nerf}
     \end{subfigure} 
\end{tabular}
\end{center}
   \caption{Visual results of (a) Mip-NeRF 360 and (b) Pre-NeRF 360: 
   our framework captures finer details in certain areas of the image, such as tree leaves, ground textures, etc and achieved more improved and better performance in just few number of iterations.
   }
   \vspace{-5mm}
\label{fig:Result_Comparison}
\end{figure}

The NeRF technique has led to significant improvements in producing realistic visual representations \cite{mildenhall2021nerf} thanks to an MLP-based approximation of the scene radiance field. However, the point-based sampling of the radiance field approximation employed by NeRFs often leads to rendering artefacts at different resolutions. To address this issue, Mip-NeRF \cite{barron2021mip} expanded upon NeRF by utilizing volumetric frustums along a cone or cylinder to reason instead. Despite the quality enhancement, NeRF and mip-NeRF encounter important challenges in unbounded scenes, where the camera can face any direction and the scene content can be at varying distances. 

Rendering extensive unbounded scenes using NeRF architectures raises three crucial challenges: \emph{parameterization}, \emph{efficiency} and \emph{ambiguity} \cite{barron2022mip}. %

\textbf{Parameterization:} refers to defining a set of parameters or constraints that can represent the space of a 3D scene. Allocating more capacity to nearby objects and less to distant ones is crucial for accurate rendering unbounded scenes. NeRFs are successful in pairing specific scene types with appropriate 3D parameterizations. Scenes unbounded in all directions require different parameterizations, as explored by NeRF++ \cite{zhang2020nerf++} and DONeRF \cite{neff2021donerf}, which shrink distant points towards the origin. Mip-NeRF 360 extends this idea to Mip-NeRF by presenting a way to apply smooth parameterization to volumes and introducing their parameterization for unbounded scenes. 

\textbf{Efficiency:} In terms of efficiency, more extensive and detailed scenes require more network capacity. Still, the process of densely querying a large MLP along each ray during training can be expensive in terms of time and resources. Therefore, NeRF and Mip-NeRF-like architectures struggle with complex and large scenes and require costly training time. Mip-NeRF 360 trained two Multi-Layer Perceptrons (MLPs) (Proposal MLP and NeRF MLP) instead of a single NeRF MLP supervised at multiple scales. The proposal MLP predicts volumetric density, which is used to resample intervals for a NeRF MLP to render the image. The Proposal MLP is not supervised by the input image, but rather by the histogram weights generated by the NeRF MLP. This approach enables us to leverage the advantages of a small proposal MLP frequently evaluated alongside a large NeRF MLP evaluated relatively few times. Consequently, the total capacity of Mip-NeRF 360 is significantly greater than that of Mip-NeRF ($\sim$15×), whereas the training time also slightly increases ($\sim$2×)  \cite{barron2022mip}. It also accelerates the training by 300\% with greatly improved rendering quality. Numerous attempts have been made to speed up the rendering of a trained NeRF by distilling or compressing it \cite{hedman2021baking, reiser2021kilonerf, yu2021plenoctrees}, but these methods do not affect the training process. Even though an octree  acceleration structure \cite{samet1990design} is constructed while optimizing a NeRF-like model in the Neural Sparse Voxel Fields \cite{liu2020neural} method, it does not result in a significant reduction in training time \cite{barron2022mip}.

\textbf{Ambiguity:} One of the main limitations of NeRF-like architectures arises from the difficulty of creating a 3D model from multiple 2D images and generating high-quality views from novel camera angles. There are many NeRFs that explain away the input images, but only a few approaches succeed in producing satisfactory outcomes for novel views \cite{ hedman2021baking, mildenhall2021nerf, oechsle2021unisurf, zhang2021nerfactor}. However, these solutions tackle different problems than Mip-NeRF 360, such as non-smooth surfaces and slow rendering \cite{barron2022mip}. Moreover, these constraints are created explicitly for the point samples that NeRF utilizes. In contrast, Mip-NeRF 360 is intended to operate with the continuous weights specified along every Mip-NeRF ray. Unfortunately, several NeRF-based techniques are even incapable of processing unbounded 360-degree scenes. For instance, RegNeRF \cite{niemeyer2022regnerf} cannot render such scenes correctly, resulting in empty or dark-brown frames since these NeRF-like architectures are not designed to handle unbounded spherical scenes.

\label{sec:reduction_over_recovery}

To eliminate ambiguity in our proposed data model, we introduce a {\em Reduction Detection layer} that filters out all the input images that are blurry, noisy or redundant. Choosing to remove the corrupted frames instead of recovering them is related to handling real-world scenes, where our methodology aims to determine the high amount of information as lowest as the number of frames. However, incorporating methods like Deblurring \cite{sun2021mefnet, chen2022simple}, Image-Super Resolution \cite{wang2021real} or Image Restoration \cite{liang2022recurrent, liang2022vrt} may have adverse effects on the resilience of our approach. When reconstructing the frame, various obstacles may arise, such as evident rectangular artifacts in the resulting frames, the omission of critical small details in the scene, the accumulation of error percentages in these methods into the final NeRF-like model, memory limitations, and most significantly, a negative impact on the system's overall speed. Additionally, these methods may rely heavily on previous knowledge to enhance images, which imposes further constraints on NeRF-like approaches.

Our Reduction Detector layer uses two filters to detect Defocus Blur and near-Image Similarity in 2D input images. Fast Fourier Transformation (FFT) \cite{aravinth2022implementation} is used with the Laplacian method to handle Defocus Blur. Whereas for near-Image Similarity, the Perceptual Hashing (P-Hash)  \cite{idealods2019imagededup}  algorithm is used with Hamming distance thresholding. These techniques help to improve the input data's accuracy and reliability by identifying and removing redundant, blurry, noisy images. These filtered and clean input images are then passed to Camera Pose Estimation (CPE) to extract more accurate and precise poses and features of the selected unbounded scene. 

\textbf{Camera Pose Estimation} is essential to feed NeRF with the viewing direction $(\theta,\Phi)$.  Camera Poses Estimation is a computer vision technique that determines the camera's position and orientation in 3D space relative to a scene. It is essential for various applications such as augmented reality \cite{baker2023localization}, 3D reconstruction \cite{feng2023predrecon}, and robotics \cite{shim2023remote}. Accurate camera pose estimation is critical for generating photo-realistic rendering of 3D scenes from NeRF. Several techniques are available, including Colmap \cite{schonberger2016structure}, SuperGlue \cite{sarlin2020superglue}, Hierarchical localization (Hloc) \cite{sarlin2019coarse}, and Pixel-Perfect Structure-from-Motion (PixSfM) \cite{lindenberger2021pixel}, which can accurately estimate the camera poses even in challenging scenarios. All NeRF-like methods used Colmap \cite{schonberger2016structure} as a preprocessing step for estimating the viewing direction $(\theta,\Phi)$. Colmap is a modern Structure from Motion (SfM) technique that uses various advanced strategies to improve reconstruction. These strategies include geometric verification, next-best view selection, robust triangulation, iterative outlier filtering, and efficient bundle adjustment parameterization via redundant view mining. Colmap is more powerful and complete than current SfM techniques \cite{schonberger2016structure} while maintaining efficiency. Generally, Structure from Motion (SfM) fails when the scene exhibits symmetries and duplicated structures \cite{cui2015global, yan2017distinguishing}, where it also applied on Colmap, as shown in Fig. \ref{fig:colmap_poses_dish}. However, prior works \cite{lindenberger2021pixel, sarlin2019coarse, sarlin2020superglue} aim to fix this problem where they still use Colmap as a backbone. In our proposed framework, we propose to integrate into the NeRF framework the Pixel-Perfect Structure-from-Motion (PixSfM) \cite{lindenberger2021pixel} as it addresses the complex scene appearance issue with high precision and shows a reasonable speed and adaptive solution to cope with memory limitation.

\textbf{PixSfM} is an advanced computer vision framework that uses innovative refinement techniques to improve the accuracy of SfM by aligning low-level image information from multiple views. It leverages the capabilities of several robust networks and is built upon the solid foundation of Hloc \cite{sarlin2019coarse} and SuperGlue \cite{sarlin2020superglue} architectures, with Colmap as a base. PixSfM improves the accuracy of camera poses and scene geometry by enhancing the initial key-point locations and refining points, and camera poses as post-processing. It is highly robust, even in challenging scenarios, and is a versatile tool for enhancing the accuracy of SfM. Therefore, in our proposed data model, we replaced Colmap with the more advanced and versatile CPE technique, PixSfM.

In summary, in this paper, we designed a framework to address the challenges faced by NeRF-like methods in generating realistic renderings of unbounded scenes, as shown in Fig. \ref{fig:Result_Comparison}. The main {\bf contributions} of this article are:
\begin{enumerate}
    \item 
    We introduce a new Reduction deduction layer in the Pre-NeRF 360 framework to address the ambiguity issue that arises in unbounded scenes. This layer incorporates two optimized filters, Defocus Blur and near-Image Similarity eliminating  noisy, blurry and redundant input images.
    \item 
    We adopt an advanced and innovative camera pose estimation technique called PixSfm \cite{lindenberger2021pixel} instead of the usually used Colmap tool for Camera Pose Estimation \cite{schonberger2016structure}. PixSfm enables us to obtain more intricate and accurate camera pose and features, which helps any NeRF-like architectures to generate highly detailed and photo realistic renderings of unbounded scenes. We show that our framework based on PixSfm is able  to extract dynamic and high-level features,  and differentiate between extracted camera positions that correspond to common features, reducing significantly the ambiguity in camera pose estimation.
    \item We offer an enhanced and upgraded version of the public Nutrition5k dataset called N5k360. We enabled it to be accepted for all NeRF-like architecture in order to create 3D view and rendering of the dishes. The N5k360 dataset contains all Nutrition5k dishes, and will be done publicly available after publishing the paper to be utilized with any NeRF-like architecture.
    \item We conducted a comprehensive set of experiments using the Mip-NeRF 360 and our proposed N5k360 dataset. In comparison with Mip-NeRF 360 dataset, our proposed framework achieved slightly improved and better performance with very fewer iterations. 
    We validated the results using three evaluation metrics (PSNR, SSIM, LPIPS) improving the state of the art performance.
\end{enumerate}

\begin{figure*}
\begin{center}
\includegraphics[width=1\linewidth, height=7cm]{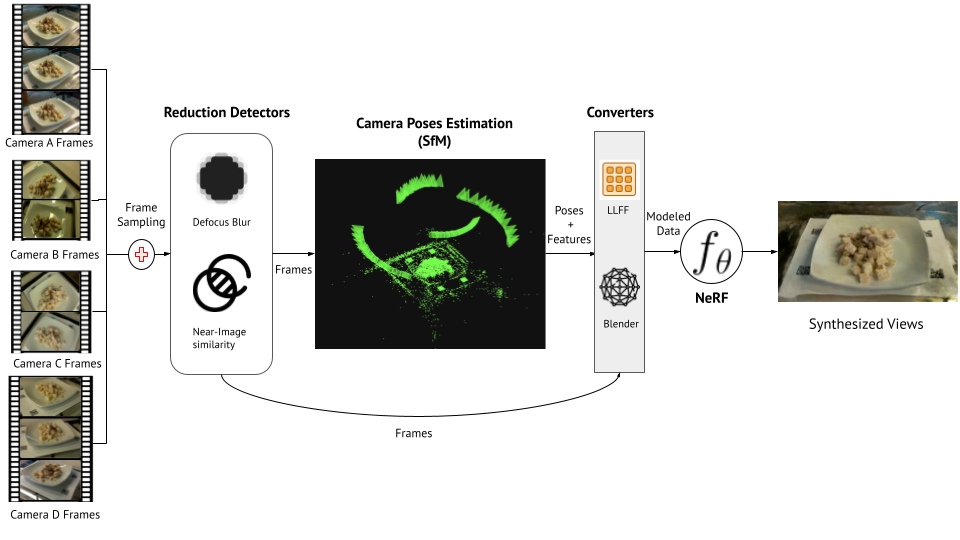}
\end{center}
   \caption{Our proposed framework diagram, which  outlines the entire workflow from a set of videos to any NeRF-like application. The data representation used in the diagram consists of four input videos that were taken from the Nutrition5k dataset.}
\label{fig:short}
\end{figure*}

\section{Proposed Methodology}
This section presents a detailed description of our framework. Our study focuses on how to improve the NeRF input data  (blurred, noisy, and redundant input images and pose estimations) in order to alleviate the ambiguity problem of NeRF-based models. 

\subsection{Overview}
\label{sec:overview}
Fig.~\ref{fig:short} shows a diagram representing the general overview of our proposed approach. The input of our model is a set of videos (or a collection of images) that is used to extract a set of key frames represented by a subsampled set of data (e.g.  every \(k^{th}\) frame) from each video. Our framework is composed of three main parts: reduction detector, camera pose estimation and converter. Within the reduction detector part, \textbf{Defocus Blur reduction} module removes all the blurry frames from the list of sampled frames; followed by a \textbf{near-Image Similarity reduction} module removing the duplicate frames from the blurry-free sampled frames. Consequently, a \textbf{Camera Poses Estimation} module extracts the low detailed features and match them, resulting in the cameras' locations (i.e. poses). Lastly, a \textbf{Converter} module reformats the Camera poses information and the frames into parsable NeRF-based formats such as LLFF or Blender.

\subsection{Preliminaries: NeRF for  View Synthesis}
\textbf{NeRFs} \cite{mildenhall2021nerf} can be formalized as a function $F_{\Theta}$ which takes as input a continuous 5D coordinate and yields a color and a density at that input location, such that:
\begin{equation*}\label{NeRF_equation}
    F_\Theta:(\normalx, d) \rightarrow  (c, \sigma) \, ,
\end{equation*}
where $\normalx = (x,y,z)$ is a 3D spatial location, $d=(\theta,\Phi)$ represents a spherical direction, $c=(r,g,b)$ is the color at the 5D input coordinate and $\sigma$ is the corresponding volume density.

NeRFs are typically parameterized as an MLP, which captures the 5D radiance field of a given 3D scene. This inherently volumetric information is then projected onto the image plane defined by a virtual camera to synthesize a new (unseen) photo-realistic image. The typical approach uses direct volume rendering, where virtual rays $r$ with origin $o$ at the virtual camera position are cast towards the scene. The number of rays equals the number of pixels in the synthetic image. Their direction $d$ is set so that each ray $r(t) = o + td$ passes through the center of each image pixel, with $t \geq 0$ being the ray parameter determining the current position along the given ray. The color of each pixel is then computed by marching along its corresponding ray and collecting the color $c$ and volume density $\sigma$ values at a set of discrete locations. Finally, the collected values are used to compose the final pixel color. This process can be formalized as \cite{mildenhall2021nerf}:
\begin{equation*}
\begin{multlined}
   \hat{C}(r) = \sum_{i=1}^{N} T_i (1-exp(-\sigma_i \delta_i)))c_i, \\
\text{with} \;\; T_i = exp \left ( -\sum_{j=1}^{i-1}\sigma_j\delta_j \right ),
\end{multlined}
\end{equation*}
where \(N\) refers to the number of sampled points across the ray, \(\delta_i = t_{i + 1} - t_i \) refers to the distance between two adjacent samples, \(c_i\) and \(\sigma_i\) refer to the per-point radiance and density, and \(T_i\) refers to the accumulated transmittance.

Since NeRF is differentiable, it is possible to define a loss function that allows NeRF training to minimize the mean-squared error (MSE) between the predicted renderings and the corresponding ground-truth colors:
\begin{equation*}
\mathcal{L}_{MSE} = \frac{1}{|\mathcal{R}|}\sum_{r \in \mathcal{R}} \left \|  \hat{C}(r) - C(r))\right \|^2
\end{equation*}

where \(\mathcal{R}\) refers to the batch of the randomly sampled rays that belong to one or all training images, while \(\hat{C}(r)\) and \(C(r)\) refer to the ground truth and output color of ray \(r\). Notably, this per-pixel optimization approach lacks holistic spatial understanding and makes NeRF sensitive to disturbance in pixel intensity.

\textbf{Camera Pose Estimation:} Since the input of NeRF is a continuous 5D coordinate containing a point in the space and the camera pose, CPE is used to compute the camera pose from a set of videos or frames. Assume CPE is a function called \(E\) that accept a set of frames (\(X\)):
\begin{equation*}
    E(X) = I(x_i, \theta, \Phi)
\end{equation*}
and returns \(I(x_i, \theta, \Phi)\), which is  the estimated view direction $(\theta, \Phi)$ with the associated frame, \(x_i \in X\).

 PixSfM \cite{lindenberger2021pixel} computes features by using deep CNNs. PixSfM provides direct alignment of low-level image information from multiple views: it first adjusts initial keypoint locations prior
to any geometric estimation, and subsequently refine points
and camera poses as a post-processing. In particular, PixSfM adjusts both keypoints and bundles, before and after reconstruction, by direct image alignment in a learned feature space. Exploiting this locally-dense information is significantly more accurate than geometric optimization, while deep, high-dimensional features extracted by a CNN ensure wider convergence in challenging conditions.   The approach first refines the 2D keypoints only from tentative matches by optimizing a direct cost over dense feature maps. The second stage operates after SfM and refines 3D points and poses with a similar feature metric cost. Thus, the formulation elegantly combines globally-discriminative sparse matching with locally-accurate dense details. The refinement is robust to large detection noise and appearance changes, due to the optimization of the  feature metric error based on dense features predicted by a neural network.

\subsection{Pre-NeRF 360}
Our framework aims to improve the input of NeRF framework in order to overcome mentioned above artefacts. Let us consider as imput multiple videos \(X\) from different cameras or the same camera, as shown in Fig. \ref{fig:short}:
\begin{equation*}\label{input_representation}
    X = \left \{ X_i | i\in (1, n) \right \}
\end{equation*}
where \(X\) is a set of videos, \(X_i\) is the i-th video, where \(i \in [0, n]\), and \(n\) is the number of videos in \(X\).

\textbf{Key frame extraction:} In order to minimise the redundant frames and speed up the process we first subsample the videos namely, from each video we select every \(k^{th}\) frame obtaining \(X'=S({X}_i, k)\).
Since duplicates in frames causes instability and less robustness in the model training \cite{zheng2016improving}, the subsampling process reduces the likelihood of duplicates in the input data.

\textbf{Defocus blur reduction} \(D_{Red}\): When the inputs are corrupted, such as compressed in JPEG format or blurred due to motion, the reconstructed scenes may show apparent artefacts. Corruptions often occur during the real-world capture and preprocessing stages. It potentially seems to lead to inaccurate reconstruction. Still, the more essential and interesting question —\textit{how different corruption severity and types impact the robustness of NeRF}— is still an open field to explore. 

In order to tackle this problem, we propose to measure image sharpness in the frequency domain using the Fast Fourier Transform (FFT) with a Blur degree threshold \(h_b\) \cite{de2013image}:  
\begin{equation*}\label{defocus_blur}
    X''=B (\forall x_i \in X') = \begin{cases}
 remove, \text{ if } FFT(x_i) \leqslant h_b \\ 
 keep, \text{ otherwise. }   
\end{cases}
\end{equation*}

\textbf{Near-Image Similarity Reduction}: The frame sampling approach of the key frame generates a list of images \({X_{red}} \subseteq X\) that still could have frames redundancy. This can occur when  the camera takes longer than  \(k^{th}\) frames to move while recording. We apply a \textbf{near-image similarity} \cite{thyagharajan2021review} method to detect the duplicates among \({X}'\), especially to handle too smooth or slow camera movement behaviour. 
To detect near-image similarity, we apply Perceptual hashing function (\(P_{Hash}\)) \cite{zauner2010implementation} that generates a unique fingerprint for each frame based on its content. \(P_{Hash}\) examines the features of a frame and generates a 64-bit number fingerprint. Then, all hashes are constructed in  BKTree \cite{burkhard1973some} data structure to find "the closest" hashes. After that, we use Hamming Distance (HD) to find the \(P\) closest hashes corresponding to similare image frames.

Assume \(H\) denotes a hash function which takes one frame as a given input and return a binary string of length \(l\). Assume \(x\) indicates a particular frame in \(X'\), and \(\hat{x}\) denotes a modified version of this frame which is "perceptually similar" to \(x\). Assume \(y\) denotes a frame that is "perceptually different" from \(x\) in \(X'\). Assume \(x'\) and \(y'\) denote hash values. \({0/1}^l\) represents binary strings of length \(l\). Then the four wanted properties of a perceptual hash are identified as follows:

(\romannum{1}) Equal distribution (unpredictability) of hash values:
\begin{equation*}\label{phash_1}
P(H(x) = x') \approx \frac{1}{2^l}, \forall x' \in {0/1}^l
\end{equation*}
where $P$ stands for probability and $l$ for the length of hash code.

(\romannum{2}) Perceptually different frames \(x\) and \(y\) exhibit pairwise independence:
\begin{equation*}
P(H(x) = x' | H(y) = y') \approx P(H(x) = x'), \forall x', y' \in {0/1}^l
\end{equation*}

(\romannum{3}) Perceptually similar frames \(x\) and \(\hat{x}\) exhibit invariance:
\begin{equation*}
P(H(x) = H(\hat{x})) \approx 1
\end{equation*}

(\romannum{4}) Perceptually different frames \(x\) and \(y\) exhibit distinction:
\begin{equation*}
P(H(x) = H(y)) \approx 0
\end{equation*}

Assume the HD (\(D(u, v)\)) is counting the number of index (\(i\)) differences between \(u\) and \(v\), where \(u\) and \(v\) are binary strings of length \(l\):
\begin{equation*}
D(u|H(x), v|H(y)) = |\{i: u_i \neq v_i, i = 1, ..., l\}|
\end{equation*}

Assume \(X_{red}=N({X}', h_s)\) \cite{idealods2019imagededup} is a function that finds all near-similarity frames within a threshold, \(h_s\) on the HD:
\begin{equation*}
\label{p-hash-reduction-thresholding}
\begin{multlined}
X_{red}=N(\forall x_i \in {X}', \forall y_i \in X') \\
= \{ y_i | P(H(x_i), H(y_i)) \approx 1, x_i \neq y_i \}
\end{multlined}
\end{equation*}
where it returns a set of frames duplicates  forming \(X_{red}\), \(X_{red} \subseteq X'\), and  for each \(x_i\subseteq X_{red}\) there is an \(y_i\subseteq X'\) that is a duplicate of \(x_i\). 

In order to achieve the efficient NeRF performance, we remove \(X_{red}\) from \(X''\) as follows:
\begin{equation*}\label{p-hash-reduction}
X''' =N_{Red}(\forall x_i \in {X}'') = \begin{cases}
 remove & \text{ if } x_i \in X_{red} \\ 
 keep & \text{ otherwise }   
\end{cases}
\end{equation*}
where the  frames in \(X'''\) are obtained removing all frames with a similar hash code controlled by the HD threshold \(h_s\) (frames with a smaller Hash distance than \(h_s\) are considered redundant and removed). Obtaining \(X'''\) with a distance lower than \(h_s\) leads to high frames overlapping.  Note that ensuring the scene has balanced (high enough, but not redundant) data is still an open question for NeRFs.

Note that  obtaining \(X'''\) is  exponentially impacted by the number of frames in \({X}''\); cleaning up the de-focus blurred images  before entering the near-Image Similarity Reduction step leads to significantly lower computation and better reduction speed.

\subsection{N5k360}

\begin{figure}[t]
\begin{center}
\includegraphics[width=1.0\linewidth, height=7cm]{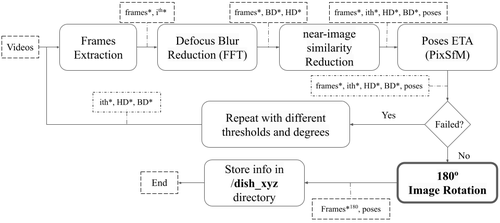}
\end{center}
   \caption{Our framework with an additional step of data rotation \(180^o\), to apply it on the  Nutrition5k dish videos. 
   }
\label{fig:workflow_n5k}
\end{figure}

In this article, we propose a new and advanced 360 food dataset named N5k360. N5k360 is an improved and enhanced version of the Nutrition5k dataset that is compatible with all NeRF architectures. Our aim was to construct  a 360º viewed 5,000 realistic food dishes that are created from Nutrition5k dataset, as mentioned in fig \ref{fig:workflow_n5k}.

 The Nutrition5k dataset contains visual and nutritional information for 5,000 realistic food dishes. These dishes were captured from Google cafeterias using a custom scanning rig. The scanning rig is equipped with four Raspberry Pi cameras (A, B, C, and D) strategically positioned to capture a 360-degree view of the dishes. Cameras A and D are mounted opposite each other at the top of the rig, while cameras B and C are positioned on the sides to cover the remaining areas of the dish.

\begin{figure}[t]
\setlength{\tabcolsep}{1pt}
\begin{center}
\begin{tabular}{ccc}
    \begin{subfigure}[b]{.3\linewidth}
         \centering
         \includegraphics[width=\textwidth]{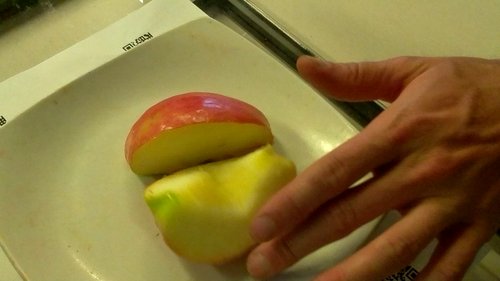}
         \setcounter{subfigure}{0}%
         \caption{Hand}
         \label{fig:hand}
     \end{subfigure} & 
     \begin{subfigure}[b]{.3\linewidth}
         \centering
         \includegraphics[width=\textwidth]{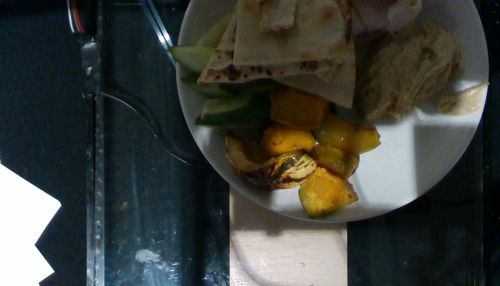}
         \setcounter{subfigure}{1}%
         \caption{Incomplete}
         \label{fig:dish_1550710793}
     \end{subfigure} &
     \begin{subfigure}[b]{.3\linewidth}
         \centering
         \includegraphics[width=\textwidth]{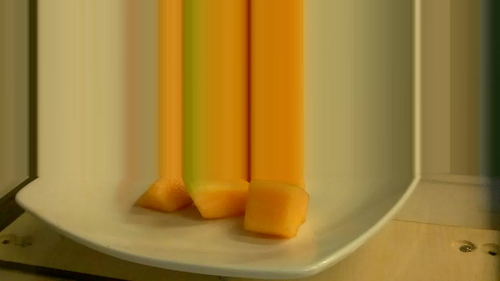}
         \setcounter{subfigure}{2}%
         \caption{Lag}
         \label{fig:lag}
     \end{subfigure} \\ 
     \begin{subfigure}[b]{.3\linewidth}
         \centering
         \includegraphics[width=\textwidth]{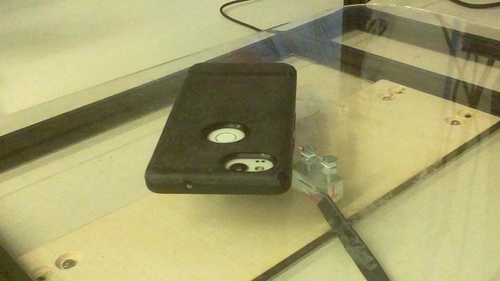}
         \setcounter{subfigure}{3}%
         \caption{Mobile}
         \label{fig:mobile}
     \end{subfigure} &
     \begin{subfigure}[b]{.3\linewidth}
         \centering
         \includegraphics[width=\textwidth]{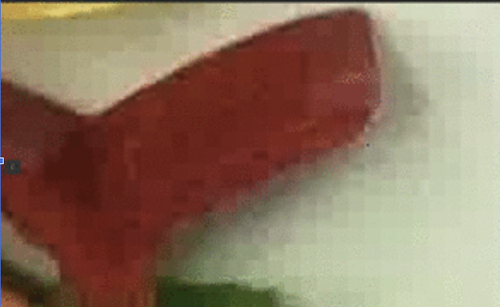}
         \setcounter{subfigure}{4}%
         \caption{Pixelated}
         \label{fig:pixelated}
     \end{subfigure} &
     \begin{subfigure}[b]{.3\linewidth}
         \centering
         \includegraphics[width=\textwidth]{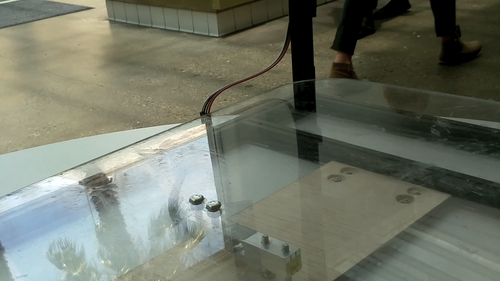}
         \setcounter{subfigure}{5}%
         \caption{Empty}
         \label{fig:empty}
     \end{subfigure} 
\end{tabular}
\end{center}
   \caption{Examples of problematic and challenging issues occured while dealing with Nutrition5k dataset.}
\label{fig:N5K_Challenges}
\end{figure}

Being Nutrition5K a real dataset refering to a complex real domain i.e. food domain, we encountered numerous issues and challenges when adapting the Nutrition5k dataset for NeRF, including low-resolution videos, errors caused by humans and cameras, and blurry and redundant images, as shown in Fig.\ref{fig:N5K_Challenges}. Since the dishes were recorded using four Raspberry Pi cameras, the resulting low-resolution videos harmed the PSNR value. Furthermore, the footage contained numerous blurry and redundant images due to the camera's brief recording pauses. Moreover, the videos also included some human errors, such as hands, mobile phones, and lagging in the recorded videos. Therefore, firstly we filtered these video images from the human errors and then passed them to our proposed data model where we applied the reduction detection layer, which filters the blurry, noisy and redundant images followed by the CPE and NeRF.

In order to generate the N5k360 dataset, we extend our framework by including an image-rotation pipeline designed to rotate inverted frames by \(180^o\) degrees. The purpose of this pipeline is to ensure proper synchronization of all frames captured by the recording cameras, as shown in Figure \ref{fig:workflow_n5k}. In some cases, it was necessary to repeat the workflow with a higher threshold \(h_b\) as the CPE process led to undesired outcomes, such as blurry frames. Insufficient features extracted from the frames caused the CPE to fail in estimating camera poses for all frames in a scene, leading to exclusion from the feature matching by the SfM.

\begin{figure*}[t]
     \centering\setlength{\tabcolsep}{1pt}
\begin{tabular}{ll|ll}
\multicolumn{2}{c|}{N5k360 Dataset} & \multicolumn{2}{c}{Mip-NeRF 360 \cite{barron2022mip} dataset} \\
 \begin{subfigure}[b]{.25\linewidth}
         \centering
         \includegraphics[width=\textwidth]{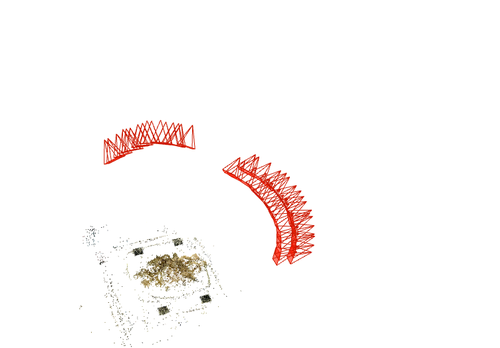}
         \setcounter{subfigure}{0}%
         \caption{ }
         \label{fig:colmap_poses_dish}
     \end{subfigure} & 
     \begin{subfigure}[b]{.25\linewidth}
         \centering
         \includegraphics[width=\textwidth]{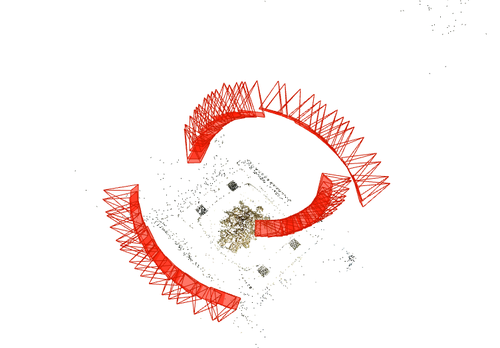}
         \setcounter{subfigure}{1}%
         \caption{ }
         \label{fig:pixsfm_poses_dish}
     \end{subfigure} &
     \begin{subfigure}[b]{.25\linewidth}
         \centering
         \includegraphics[width=\textwidth]{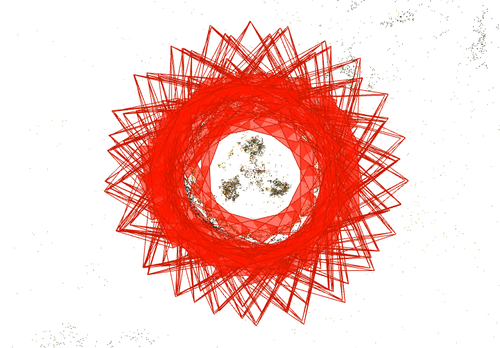}
         \setcounter{subfigure}{2}%
         \caption{ }
         \label{fig:colmap_poses}
     \end{subfigure} & 
     \begin{subfigure}[b]{.25\linewidth}
         \centering
         \includegraphics[width=\textwidth]{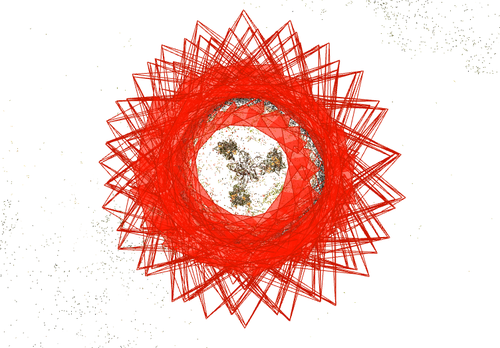}
         \setcounter{subfigure}{3}%
         \caption{ }
         \label{fig:pixsfm_poses}
     \end{subfigure}
     
\end{tabular}
        \caption{Comparison between Colmap \& PixSfM feature extractions: (a) The extracted features of Colmap using the N5k360 dataset. (b) The extracted features of PixSfM using N5k360 dataset. (c) The extracted features of Colmap using Mip-NeRF 360 dataset. (d) The extracted features of PixSfM using Mip-NeRF 360 dataset.}
        \label{fig:workflow_n5k_camera_poses_comparaison}
\end{figure*}

\section{Experimental Results}
\label{sec:results}
In this section we comment the Mip-NeRF 360 dataset, the implementation setting, the results and the comparison of PixSfM to Colmap CPE. Our proposed framework was evaluated on both multifaceted datasets, Mip-NeRF 360 and N5k360. As usual for NeRF evaluation, we provide the results for three error metrics, namely PSNR, SSIM and LPIPS. The Training Time (hrs), Model Parameters (M), number of Iterations and the Equivalent number of Iterations to TPU are also presented for all the scenes in the datasets. 

\subsection{Dataset}
\begin{figure}[t]
\setlength{\tabcolsep}{1pt}

\begin{center}
\begin{tabular}{cccc}
\multicolumn{4}{c}{Indoor} \\ 
     \begin{subfigure}[b]{.25\linewidth}
         \centering
         \includegraphics[width=\textwidth]{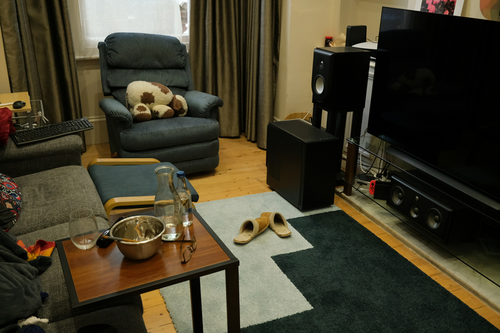}
         \setcounter{subfigure}{0}%
         \caption{Room}
         \label{fig:360Room}
     \end{subfigure} & 
     \begin{subfigure}[b]{.25\linewidth}
         \centering
         \includegraphics[width=\textwidth]{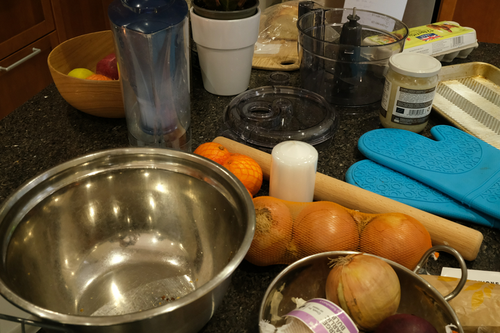}
         \setcounter{subfigure}{1}%
         \caption{Counter}
         \label{fig:360counter}
     \end{subfigure} &
     \begin{subfigure}[b]{.25\linewidth}
         \centering
         \includegraphics[width=\textwidth]{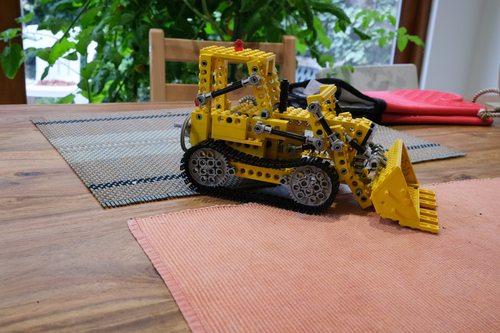}
         \setcounter{subfigure}{2}%
         \caption{Kitchen}
         \label{fig:360kitchen}
     \end{subfigure} &
     \begin{subfigure}[b]{.25\linewidth}
         \centering
         \includegraphics[width=\textwidth]{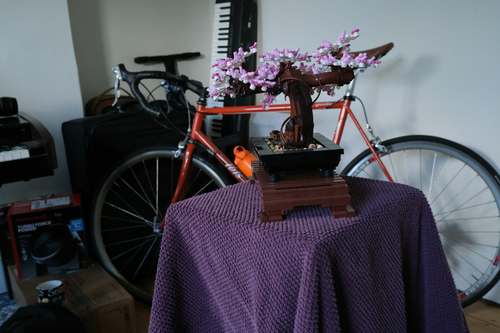}
         \setcounter{subfigure}{3}%
         \caption{Bonsai}
         \label{fig:360Bonsai}
     \end{subfigure}
\end{tabular}

\begin{tabular}{ccccc}
    \multicolumn{5}{c}{Outdoor} \\
    \begin{subfigure}[b]{.2\linewidth}
         \centering
         \includegraphics[width=\textwidth]{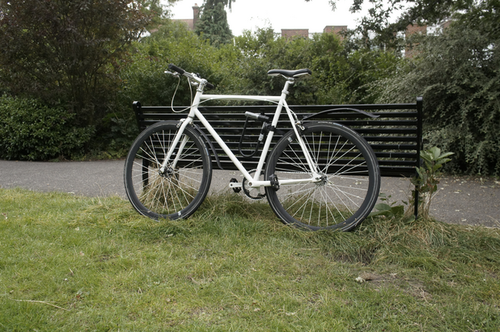}
         \setcounter{subfigure}{4}%
         \caption{Bicycle}
         \label{fig:360bicycle}
     \end{subfigure} & 
     \begin{subfigure}[b]{.2\linewidth}
         \centering
         \includegraphics[width=\textwidth]{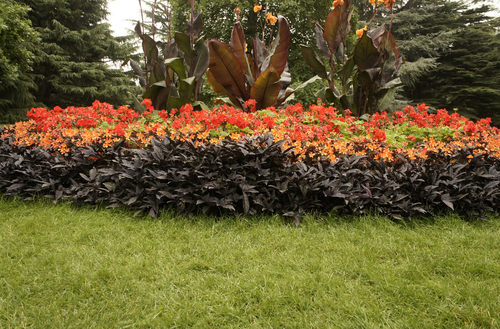}
         \setcounter{subfigure}{5}%
         \caption{Flowers}
         \label{fig:360Flowers}
     \end{subfigure} &
     \begin{subfigure}[b]{.2\linewidth}
         \centering
         \includegraphics[width=\textwidth]{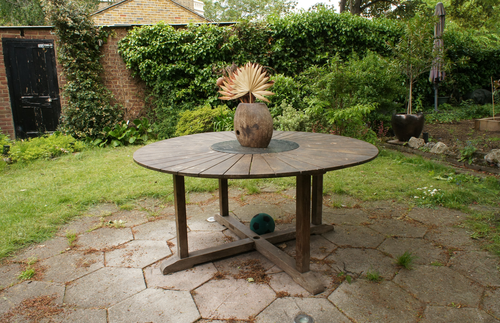}
         \setcounter{subfigure}{6}%
         \caption{Garden}
         \label{fig:360Garden}
     \end{subfigure} & 
     \begin{subfigure}[b]{.2\linewidth}
         \centering
         \includegraphics[width=\textwidth]{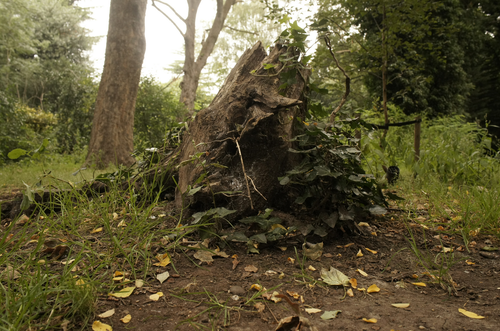}
         \setcounter{subfigure}{7}%
         \caption{Stump}
         \label{fig:360Stump}
     \end{subfigure}  &
    \begin{subfigure}[b]{.2\linewidth}
         \centering
         \includegraphics[width=\textwidth]{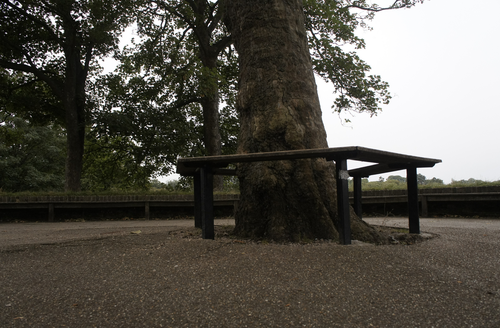}
         \setcounter{subfigure}{8}%
         \caption{Treehill}
         \label{fig:360treehill}
    \end{subfigure}
\end{tabular}

\end{center}
   \caption{Examples of pictures from the Mip-NeRF 360 \cite{barron2022mip} dataset, pictures Figs. (~\ref{fig:360bicycle}-~\ref{fig:360treehill}) are representing Outdoor scenes showing much more challenging scenes for the NeRF. In contrast, Figs. (~\ref{fig:360Room}-~\ref{fig:360Bonsai}) are representing the Indoor scenes, which show lower degree challenges than Outdoor scenes.}
\label{fig:example_datasets}
\end{figure}

The Mip-NeRF 360 dataset \cite{barron2022mip} comprises 9 different scenes, including 5 outdoor and 4 indoor scenes. Each scene features a sophisticated central area and intricates background details. To capture these scenes, specific measures were taken to minimize photometric variations by fixing camera exposure settings, minimizing lighting changes and avoiding moving objects. The examples of these scenes are illustrated in Fig.\ref{fig:example_datasets}.

\begin{figure}[t]
\setlength{\tabcolsep}{1pt}
\begin{center}
\begin{tabular}{cccc}
    \begin{subfigure}[b]{.25\linewidth}
         \centering
         \includegraphics[width=\textwidth]{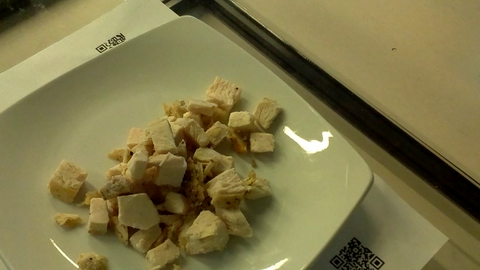}
         \setcounter{subfigure}{0}%
         \caption{ }
         \label{fig:dish_1550704750}
     \end{subfigure} & 
     \begin{subfigure}[b]{.25\linewidth}
         \centering
         \includegraphics[width=\textwidth]{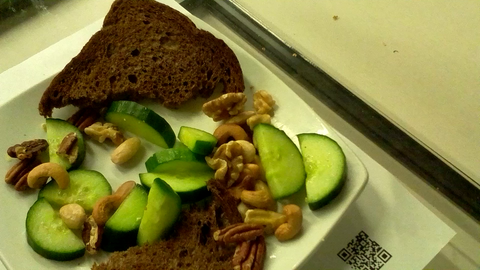}
         \setcounter{subfigure}{1}%
         \caption{ }
         \label{fig:dish_1550710793}
     \end{subfigure} &
     \begin{subfigure}[b]{.25\linewidth}
         \centering
         \includegraphics[width=\textwidth]{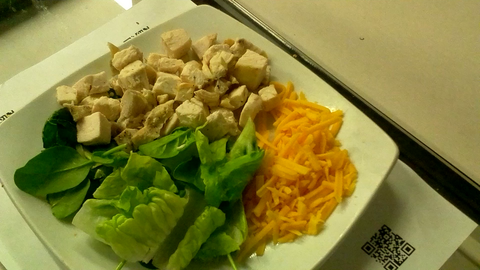}
         \setcounter{subfigure}{2}%
         \caption{ }
         \label{fig:dish_1550712459}
     \end{subfigure} & 
     \begin{subfigure}[b]{.25\linewidth}
         \centering
         \includegraphics[width=\textwidth]{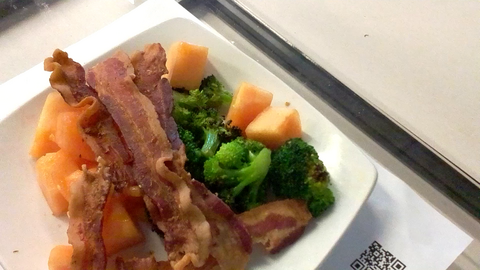}
         \setcounter{subfigure}{3}%
         \caption{ }
         \label{fig:dish_1550712459}
     \end{subfigure} \\
     \begin{subfigure}[b]{.25\linewidth}
         \centering
         \includegraphics[width=\textwidth]{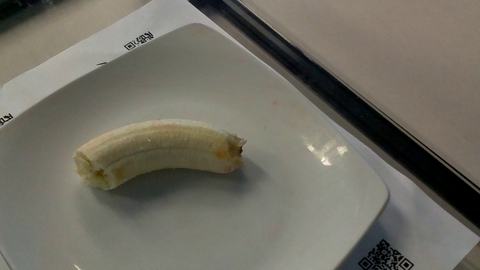}
         \setcounter{subfigure}{4}%
         \caption{ }
         \label{fig:dish_1550775219}
     \end{subfigure} &
     \begin{subfigure}[b]{.25\linewidth}
         \centering
         \includegraphics[width=\textwidth]{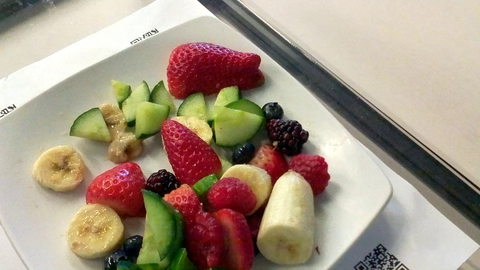}
         \setcounter{subfigure}{5}%
         \caption{ }
         \label{fig:dish_1550777256}
     \end{subfigure} &
     \begin{subfigure}[b]{.25\linewidth}
         \centering
         \includegraphics[width=\textwidth]{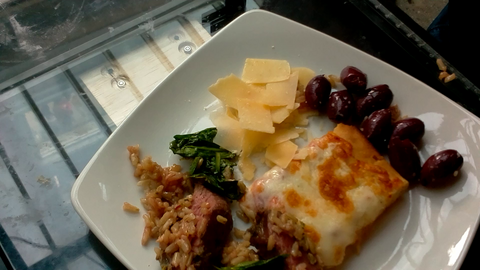}
         \setcounter{subfigure}{6}%
         \caption{ }
         \label{fig:dish_1561575996}
     \end{subfigure} &
      \begin{subfigure}[b]{.25\linewidth}
         \centering
         \includegraphics[width=\textwidth]{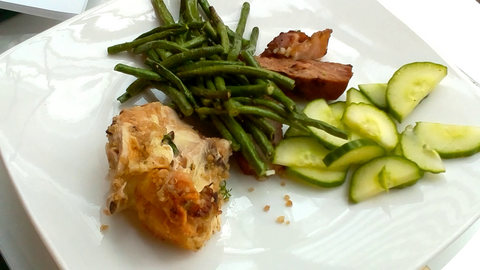}
         \setcounter{subfigure}{7}%
         \caption{ }
         \label{fig:dish_1561664061}
     \end{subfigure}
\end{tabular}
\end{center}
   \caption{Examples of pictures from  N5k360 representing different Nutrition dishes.}
\label{fig:n5k_example_datasets}
\end{figure}

\subsection{Implementation setting}
\label{sec:resource_limitation}
We used TPU v2 with 32 cores \cite{jouppi2017datacenter}, and linear scaling rule \cite{goyal2017accurate} to fit our model configurations with NVIDIA GeForce RTX 3090/24G. Our model has \(4096\) as a batch size, a learning rate that is annealed log-linearly from \(5 \times 10^{-4}\) to \(5 \times 10^{-6}\), and  \(1 \times 10^6\) iterations. Simultaneously, we allocated 2 TPU v2 with 7 cores for each node without using the linear scaling rule for 3 days.

\subsection{Pre-NeRF 360 Results}
\label{sec:results_mipnerf_360}
We evaluate our model on the two datasets: Mip-NeRF 360 dataset and the N5k360 dataset.
In Table ~\ref{table:avg_comparasion}, we present the PSNR, SSIM \cite{wang2004image}, LPIPS \cite{zhang2018unreasonable}, Time (hours), and Iterations mean values for all NeRF-like methods for the nine scenes. One can note that our model outperforms all NeRF-like methods by ~1.18x in LPIPS. Note that our model needs 3.17x less number of iterations. Moreover, our model outperforms Mip-NeRF 360 \cite{barron2022mip} w/GLO by ~2.59\%, while Mip-NeRF ~2.64\% in PSNR. In contrast, our model outperforms all NeRF-like methods, except Mip-NeRF 360 and Mip-NeRF 360 w/GLO in SSIM outperforming our model by 2.02\% and 1.27\%, accordingly. 
\begin{table}[t]
\setlength{\tabcolsep}{1pt}
\begin{center}
\begin{tabular}{|l|ccc|c|c|c|}
\hline
    & PSNR↑ & SSIM↑ & LPIPS↓ & Time(hrs) & Params & Iters \\
\hline\hline
\cite{dengjaxnerf, mildenhall2021nerf} & 23.85  & 0.605  & 0.451   & \cellcolor{orange!25} 4.16       & \cellcolor{yellow!25} 1.5M      & 250k       \\
\cite{neff2021donerf}        & 24.03  & 0.607  & 0.455   & \cellcolor{yellow!25} 4.59       & \cellcolor{orange!25} 1.4M      & 250k       \\
\cite{barron2021mip}                 & 24.04  & 0.616  & 0.441   & \cellcolor{red!25} 3.17       & \cellcolor{red!25} 0.7M      & 250k       \\
\cite{zhang2020nerf++}                   & 25.11  & 0.676  & 0.375   & 9.45       & 2.4M      & 250k       \\
\cite{hedman2018deep}            & 23.70  & 0.666  & 0.318   & -          & -         & 250k       \\
\cite{kopanas2021point} & 23.71  & 0.735  & 0.252   & -          & -         & 250k       \\
\cite{riegler2021stable}     & 25.33  & 0.771  & \cellcolor{yellow!25} 0.211   & -          & -         & 250k       \\
\cite{barron2021mip}*         & 26.19  & 0.748  & 0.285   & 22.71      & 9.0M      & 250k       \\
\cite{zhang2020nerf++}*         & 26.39  & 0.750  & 0.293   & 19.88      & 9.0M      & 250k       \\
\cite{barron2022mip}                          & \cellcolor{red!25} 27.69  & \cellcolor{red!25} 0.792  & \cellcolor{orange!25} 0.237   &  6.89       & 9.9M      & 250k       \\
\cite{barron2022mip}+            & \cellcolor{yellow!25} 26.26 & \cellcolor{orange!25} 0.786 & \cellcolor{orange!25} 0.237 & 6.90 & 9.9M & 250k \\
 
Ours                             & \cellcolor{orange!25} 26.96  & \cellcolor{yellow!25} 0.776  & \cellcolor{red!25}0.201   & 15.80      & 9.0M      & \cellcolor{red!25}78,75k \\
\hline
\end{tabular}
\end{center}
\caption{A quantitative comparison of our model with the SOTA on the Mip-NeRF 360 dataset. (*) denotes the NeRF-like method with Bigger MLP, while (+) denotes to NeRF-like method with Generative Latent Optimization (GLO) \cite{martin2021nerf}.}
\label{table:avg_comparasion}
\end{table}

To validate our framework on the N5k360 dataset, we randomly selected the whole set of videos for 8 dishes from the Nutrition5k dataset and evaluated them on our proposed framework. These 8 dishes are shown in Fig.\ref{fig:n5k_example_datasets}, whereas the evaluations are shown in Table \ref{table:avg_comparasion_n5k360}.
We present the PSNR, SSIM \cite{wang2004image}, LPIPS \cite{zhang2018unreasonable}, Time (hours), and Iterations values for our model for  randomly selected 8 scenes from N5k360 dataset. Our model shows in average of 24.25 PSNR, 0.81 SSIM, and 0.13 LPIPS, while the training time is 55 minutes for 50k iterations. 

\begin{table}[t]
\setlength{\tabcolsep}{1pt}
\begin{center}
\begin{tabular}{|c|ccc|c|c|c|}
\hline
Dish ID & PSNR↑ & SSIM↑ & LPIPS↓ & Time(hrs) & Params & Iters \\
\hline\hline
50704750 & 25.55 & 0.85 & 0.133 & 0.55 & 9.0M & 50k \\
50710793 & 24.53 & 0.80 & 0.12 & 0.55 & 9.0M & 50k \\
50712459 & 25.57 & 0.82 & 0.11 & 0.55 & 9.0M & 50k \\
50772617 & 26.86 & 0.86 & 0.11 & 0.55 & 9.0M & 50k \\
50775219 & 26.86 & 26.86 & 0.11 & 0.55 & 9.0M & 50k \\
50777256 & 24.99 & 0.82 & 0.11 & 0.55 & 9.0M & 50k \\
61575996 & 17.54 & 0.60 & 0.22 & 0.55 & 9.0M & 50k \\
61664061 & 22.10 & 0.81 & 0.11 & 0.55 & 9.0M & 50k \\
Average & 24.25 & 0.81 & 0.13 & 0.55 & 9.0M & 50k \\
 \hline
\end{tabular}
\end{center}
\caption{A quantitative results of our model using randomly selected scenes from N5k360.}
\label{table:avg_comparasion_n5k360}
\end{table}

\subsection{PixSfM vs Colmap CPE Comparison}
\label{sec:colmap_pixsfm_comparison}
Based on our experiments, we found that the classical frame-matching NeRF paradigm for CPE, i.e. Colmap, detects key points per frame once and for all, which can yield poorly-localized features and propagate significant errors to the final geometry. Furthermore, Colmap fails to detect low-level frame information, more precisely, the keypoints from multiple views, which makes Colmap's matching process failing to find any proper matches. Moreover, it gives view directions $(\theta,\Phi)$ with a different range of error margins. For instance, when the camera symmetrically captured a low-level information indoor scene such as an N5k360 dish, the estimated camera poses are overlapped as shown in Fig. \ref{fig:workflow_n5k_camera_poses_comparaison}.a,  while PixSfM CPE can detect them precisely as shown in Fig. \ref{fig:workflow_n5k_camera_poses_comparaison}.b. 
For instance, considering the amount of  localized features, our experiments show that Colmap found fewer features as shown in  \ref{fig:workflow_n5k_camera_poses_comparaison}.c, while PixSfM CPE found much richer features.

\section{Conclusions and Future Work}
\label{sec:conclusion}
In this paper, we presented a stable data framework for volume rendering  that enhances  NeRF-like frameworks. We found that a well-preparing complex and challenging data for a NeRF-like models is error-prone and time-consuming task. Using a Reduction detection layer with Defocus blur and near image similarity detection allows to filter most NeRF challenges faced by unbounded scenes. Futhermore, we adopted a more advanced and efficient camera pose estimation technique represnted by PixSfM, enabling us to extract more detailed and precise poses from the filtered images, that is critical for the NeRF in order to assure feasible and robust results. 
Our data model outperformed the performance of the majority of NeRF-like frameworks after just a few training iterations. Additionally, we also utilized our data model to create N5k360 dataset that is the first volumetric food representation employing the proposed NeRF-based framework.
   
Our findings shed important light on the stability of NeRF-based models. Still, the more essential and interesting question —\textit{how balanced and imbalanced camera locations for a 360 scene impact the robustness of NeRF}— is still open. Moreover, more corruption reduction methods can be integrated to extend our Reduction Detector module. We hope this work can help providing a backbone and stimulating additional studies to create more reliable NeRF-like systems for real-world applications.

{\small
\bibliographystyle{ieee_fullname}
\bibliography{main}
}
\newpage\clearpage

\begin{appendices}
\newcommand{\ric}[1]{\textcolor{blue}{#1}}

\section{Additional Results}

In this appendix, we present an extra set of results and discussion which could not be included in the main paper due to space constraints. These extra results further support the superior performance of our approach with respect to the state of the art. Finally, we would like to recall that, as mentioned in Sec. \ref{sec:resource_limitation} of the main paper, the presented results were generated with a much smaller number of training iterations, compared to the state-of-the-art.

\subsection{Datasets}
 As discussed in the main paper, we validated the performance of our framework on both, Mip-NeRF 360 \cite{barron2022mip} and N5k360 datasets. Our model specializes in unbounded scenes, i.e., scenes that are not restricted by lighting and temporary obstructions constraints, as illustrated in Fig. ~\ref{fig:N5K_Challenges} in the main paper. Therefore, the datasets we considered provide complex unbounded scenes that represent a strong test for our proposal. This is in contrast with the dataset used in the ``NeRF in the Wild'' \cite{martin2021nerf} approach that targets unconstrained scenes where the scene is captured under uncontrolled settings, which causes limitations that prevent it from accurately representing numerous common, real-life occurrences in unregulated images, such as changes in lighting or temporary obstructions.

The Mip-NeRF 360 dataset is a collection of images that capture a wide range of complex scenes from various vantage points, offering unparalleled levels of detail and contextual information. On the other hand, the N5k360 dataset is food-centric, presenting a 360 view of food from four different cameras with two distinct angles. Rendering such datasets can be intricate and perplexing for NeRF-like architectures, as their camera pose estimation approach (Colmap \cite{schonberger2016structure}) fails to extract rich and correct poses from multiple videos of different camera angles, as shown in Fig \ref{fig:workflow_n5k_camera_poses_comparaison} in the main paper. Our proposed framework (Pre-NeRF 360) successfully handled both datasets and utilized their respective complexities and flexibilities to generate highly detailed and precise 360-degree renderings and depth maps for all scenes, as shown in Fig \ref{fig:three graphs}. Furthermore, Pre-NeRF 360 effectively handles a large number of input images, as demonstrated by its performance with the Mip-NeRF 360 dataset, as well as with a smaller number of input images, as exemplified by the N5k360 dataset.

\begin{table}[t]
\setlength{\tabcolsep}{1pt}
\begin{center}
\begin{tabular}{|l|c|c|c|}
\hline
Classes & \# Images & Resolution \\
\hline\hline

Bicycle  & 194  & 4946$\times$3286 \\
Bonsai   & 291  & 3118$\times$2078 \\
Counter  & 238  & 3115$\times$2076 \\
Flowers  & 173  & 5025$\times$3312 \\
Garden   & 184  & 5187$\times$3361 \\
Kitchen  & 270  & 3115$\times$2078 \\
Room     & 307  & 3114$\times$2075 \\
Stump    & 124  & 4978$\times$3300 \\
Treehill & 141  & 5068$\times$3326 \\

\hline
\end{tabular}
\end{center}
\caption{mip-NeRF 360 Dataset Description}
\label{table:mipNeRF360_dataset_description}
\end{table}

\begin{table}[t]
\setlength{\tabcolsep}{1pt}
\begin{center}
\begin{tabular}{|l|c|c|c|}
\hline
Classes & \# Images & Resolution \\
\hline\hline

Dish (50704750)  & 65  & 1920$\times$1080 \\
Dish (50710793)  & 61  & 1920$\times$1080 \\
Dish (50712459)  & 62  & 1920$\times$1080 \\
Dish (50772617)  & 59  & 1920$\times$1080 \\
Dish (50775219)  & 62  & 1920$\times$1080 \\
Dish (50777256)  & 59  & 1920$\times$1080 \\
Dish (61575996)  & 20  & 1920$\times$1080 \\
Dish (61664061)  & 22  & 1920$\times$1080 \\

\hline
\end{tabular}
\end{center}
\caption{ N5k360 dataset description.}
\label{table:n5k360_dataset}
\end{table}

 The detailed information of both datasets  regarding scene types, number of images and resolution, is presented in Table \ref{table:mipNeRF360_dataset_description} and Table \ref{table:n5k360_dataset}.
Table \ref{table:mipNeRF360_dataset_description} shows that the Mip-NeRF 360 dataset contains an average of 214 images per scene, with a maximum of 307 images for the \textit{room} indoor scene and a minimum of 124 images for the \textit{stump} outdoor scene. In contrast, Table \ref{table:n5k360_dataset} reveals that our N5k360 dataset contains an average of 52 images per scene, with a maximum of 65 images for the \textit{Dish (50704750)} scene and a minimum of 20 images for the \textit{Dish (61575996)} scene. Interestingly,  the average number of images in the Mip-NeRF 360 dataset is 214\% larger than that of our N5k360 dataset. Furthermore, the average resolution of a scene in the Mip-NeRF 360 dataset is 4185$\times$2766, while our N5k360 dataset has an average resolution of 1920$\times$1080, meaning that the Mip-NeRF 360 dataset has a resolution that is 5.88$\times$ higher than that of our eight scenes in the N5k360 dataset.

The main objective of creating the N5k360 dataset is to establish a comprehensive food dataset with a \(360^o\) view, that can maximize the utility of NeRF-based techniques in confronting a variety of food-related challenges, such as food volume estimation \cite{amugongo2023mobile}, Food 3D Reconstruction \cite{dalai2023accurate}, and Food as medicine \cite{pieroni2006eating} etc. To demonstrate the effectiveness of our approach, we randomly selected eight \(360^o\) nutrition dishes and used them to train our Mip-NeRF 360 model. Detailed quantitative and qualitative evaluations are presented in Sec.~\ref{sec:results}.

\subsection{Qualitative and quantitative results}

\begin{figure*}
     \centering\setlength{\tabcolsep}{2pt}
     \captionsetup{font=small, justification=centering, singlelinecheck=false}
\begin{tabular}{lll}
 \begin{subfigure}[b]{.3\linewidth}
         \centering
         \includegraphics[width=\textwidth]{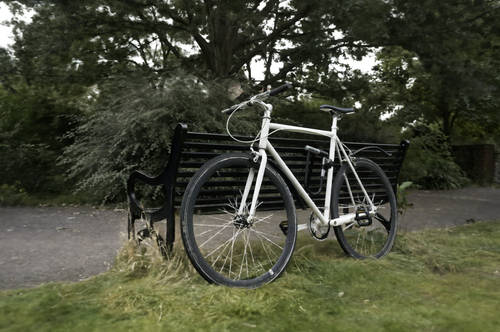}
         \setcounter{subfigure}{0}%
         \caption{Bicycle\\PSNR=26.302, SSIM=0.670, LPIPS=0.334}
         \label{fig:bicycle}
     \end{subfigure} & 
     \begin{subfigure}[b]{.3\linewidth}
         \centering
         \includegraphics[width=\textwidth]{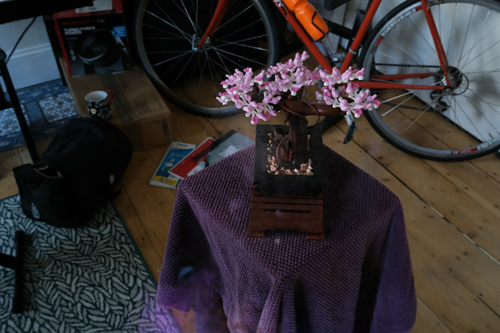}
         \setcounter{subfigure}{1}%
         \caption{Bonsai\\PSNR=30.541, SSIM=0.91, LPIPS=0.036}
         \label{fig:bonsai}
     \end{subfigure} & 
     \begin{subfigure}[b]{.3\linewidth}
         \centering
         \includegraphics[width=\textwidth]{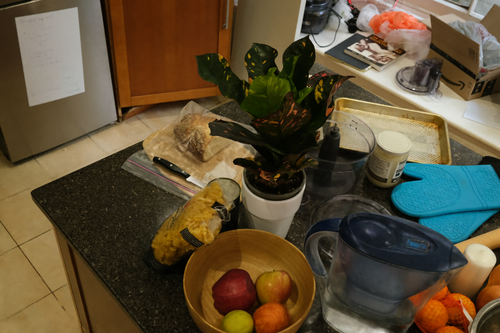}
         \setcounter{subfigure}{2}%
         \caption{Counter\\PSNR=29.410, SSIM=0.905, LPIPS=0.079}
         \label{fig:counter}
     \end{subfigure} \\

 \begin{subfigure}[b]{.3\linewidth}
         \centering
         \includegraphics[width=\textwidth]{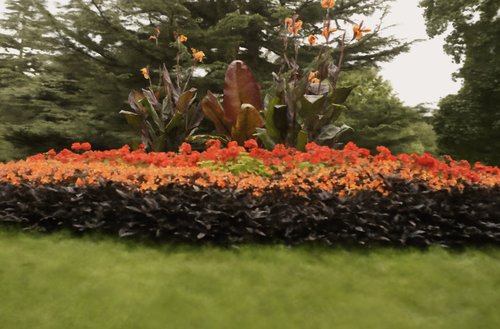}
         \setcounter{subfigure}{3}%
         \caption{Flowers\\PSNR=20.852, SSIM=0.603, LPIPS=0.350}
         \label{fig:flowers}
     \end{subfigure} & 
     \begin{subfigure}[b]{.3\linewidth}
         \centering
         \includegraphics[width=\textwidth]{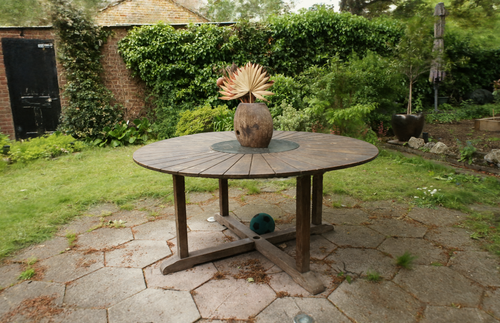}
         \setcounter{subfigure}{3}%
         \caption{Garden\\PSNR=27.25, SSIM=0.824, LPIPS=0.174}
         \label{fig:garden}
     \end{subfigure} &
      \begin{subfigure}[b]{.3\linewidth}
         \centering
         \includegraphics[width=\textwidth]{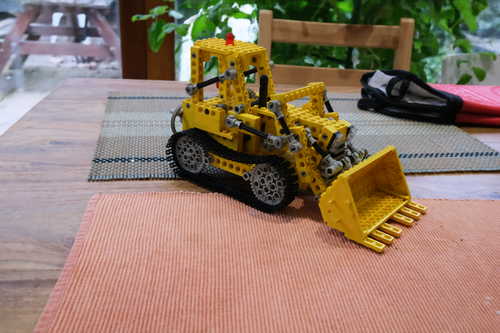}
         \setcounter{subfigure}{3}%
         \caption{Kitchen\\PSNR=31.085, SSIM=0.918, LPIPS=0.062}
         \label{fig:kitchen}
     \end{subfigure} \\
     
    \begin{subfigure}[b]{.3\linewidth}
         \centering
         \includegraphics[width=\textwidth]{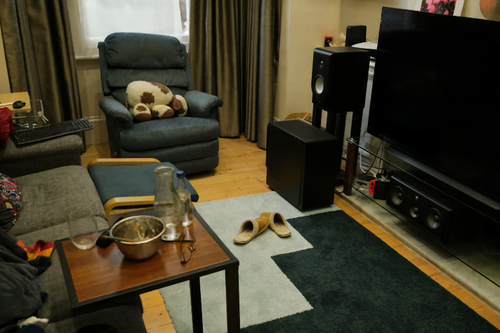}
         \setcounter{subfigure}{3}%
         \caption{Room\\PSNR=31.842, SSIM=0.930, LPIPS=0.047}
         \label{fig:room}
     \end{subfigure} & 
     \begin{subfigure}[b]{.3\linewidth}
         \centering
         \includegraphics[width=\textwidth]{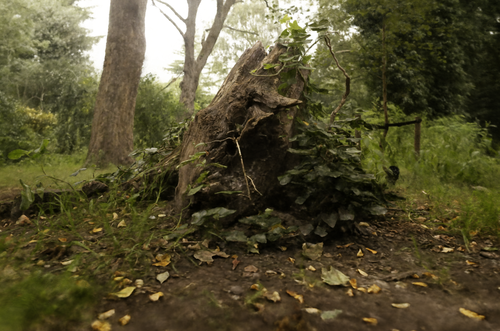}
         \setcounter{subfigure}{3}%
         \caption{Stump\\PSNR=28.619, SSIM=0.810, LPIPS=0.214}
         \label{fig:Stump}
     \end{subfigure} & 
     \begin{subfigure}[b]{.3\linewidth}
         \centering
         \includegraphics[width=\textwidth]{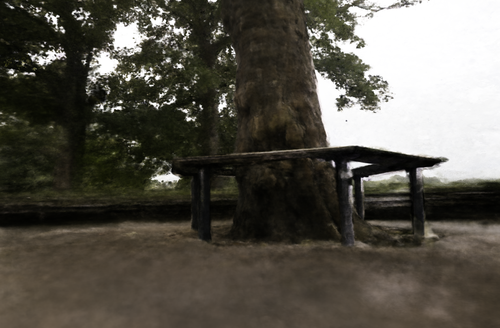}
         \setcounter{subfigure}{3}%
         \caption{Treehill\\PSNR=16.731, SSIM=0.420, LPIPS=0.511}
         \label{fig:treehill}
     \end{subfigure}
\end{tabular}
        \captionsetup{font=small, justification=raggedright, singlelinecheck=false}
        \caption{The qualitative results of our framework, including each scene's associated mean PSNR, SSIM, and LPIPS metrics. Our findings reveal that the treehill and flowers scenes were the most challenging of all the scenes.}
        \label{fig:three_graphs}
\end{figure*}

In order to emphasize the qualitative and quantitative results of both datasets, Fig \ref{fig:three_graphs} presents visual renderings of all 9 classes from the Mip-NeRF 360 dataset, including their evaluation metrics such as PSNR, SSIM, and LPIPS. Our framework shows a blurry rendering for the highly textured areas like the green grass and intensely detailed ground or leaves, as seen in Fig. \ref{fig:flowers} and Fig \ref{fig:treehill}. On the other hand, in indoor scenes, our framework outperforms all NeRF-like methods for room and counter scenes (as shown in Table \ref{table:detailed_results}), and illustrated as in Fig. \ref{fig:room} and Fig. \ref{fig:counter}, while for Kitchen (Fig. \ref{fig:kitchen}) and Bonsai (Fig. \ref{fig:bonsai}) it competes with the NeRF state-of-the-arts.

\begin{figure}
     \centering\setlength{\tabcolsep}{2pt}
\begin{tabular}{ll}
 \begin{subfigure}[b]{.5\linewidth}
         \centering
         \includegraphics[width=\textwidth]{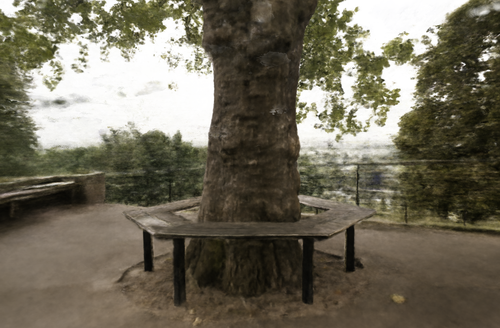}
         \setcounter{subfigure}{0}%
         \caption{Treehill rendered frame}
         \label{fig:bicycle}
     \end{subfigure} & 
     \begin{subfigure}[b]{.5\linewidth}
         \centering
         \includegraphics[width=\textwidth]{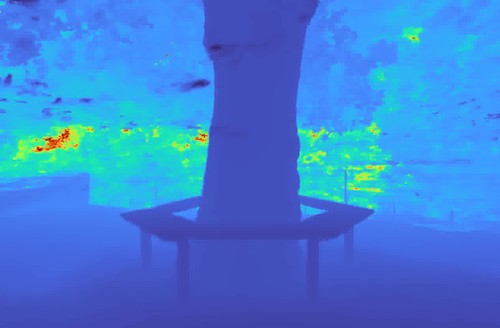}
         \setcounter{subfigure}{1}%
         \caption{Treehill depth map}
         \label{fig:bonsai}
     \end{subfigure}
     
\end{tabular}
        \caption{An example of the qualitative results produced by Pre-NeRF 360: a rendered frame accompanied by its corresponding depth map.}
        \label{fig:three graphs}
\end{figure}

\begin{figure*}
     \centering\setlength{\tabcolsep}{2pt}
     \captionsetup{justification=centering}
\begin{tabular}{ll}
 \begin{subfigure}[b]{.4\linewidth}
        
         \centering
         \includegraphics[width=\textwidth]{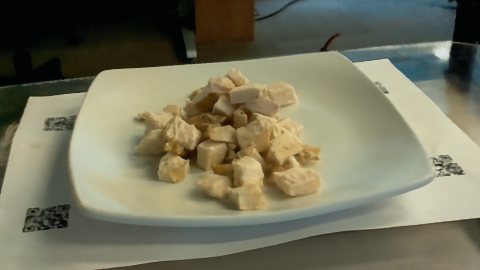}
         \setcounter{subfigure}{0}%
         \caption{Dish(50704750)\\PSNR=25.553, SSIM=0.856, LPIPS=0.133}
         \label{fig:r_dish_1550704750}
     \end{subfigure} & 
     \begin{subfigure}[b]{.4\linewidth}
         \centering
         \includegraphics[width=\textwidth]{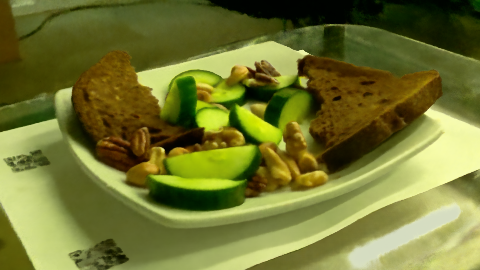}
         \setcounter{subfigure}{1}%
         \caption{Dish(50710793)\\PSNR=24.532, SSIM=0.780, LPIPS=0.126}
         \label{fig:r_dish_1550710793}
     \end{subfigure} \\
     \begin{subfigure}[b]{.4\linewidth}
         \centering
         \includegraphics[width=\textwidth]{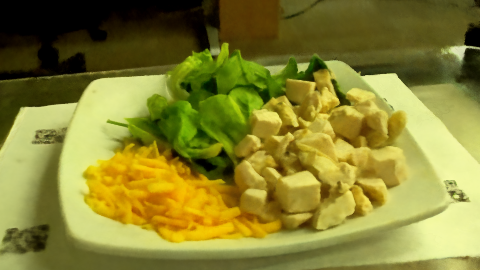}
         \setcounter{subfigure}{2}%
         \caption{Dish(50712459)\\PSNR=25.574, SSIM=0.828, LPIPS=0.116}
         \label{fig:r_dish_1550712459}
     \end{subfigure} &
      \begin{subfigure}[b]{.4\linewidth}
         \centering
         \includegraphics[width=\textwidth]{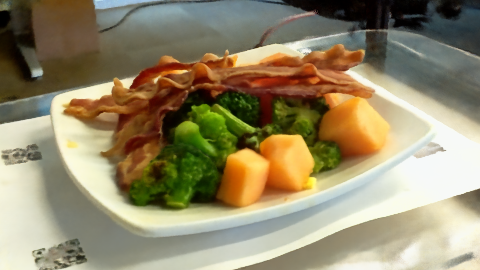}
         \setcounter{subfigure}{3}%
         \caption{Dish(50772617)\\PSNR=26.864, SSIM=0.862, LPIPS=0.116}
         \label{fig:r_dish_1550772617}
     \end{subfigure} \\
      \begin{subfigure}[b]{.4\linewidth}
         \centering
         \includegraphics[width=\textwidth]{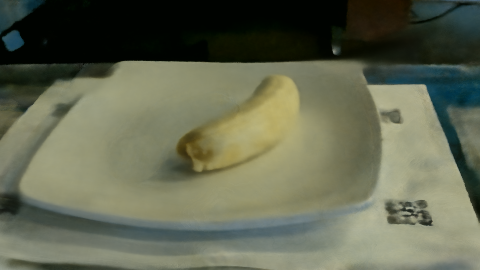}
         \setcounter{subfigure}{4}%
         \caption{Dish(50775219)\\PSNR=26.865, SSIM=0.862, LPIPS=0.116}
         \label{fig:r_dish_1550775219}
     \end{subfigure} &
      \begin{subfigure}[b]{.4\linewidth}
         \centering
         \includegraphics[width=\textwidth]{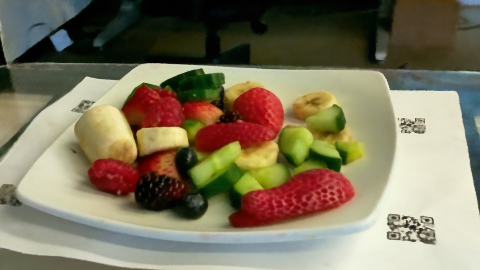}
         \setcounter{subfigure}{5}%
         \caption{Dish(50777256)\\PSNR=24.999, SSIM=0.829, LPIPS=0.111}
         \label{fig:r_dish_1550777256}
     \end{subfigure} \\
      \begin{subfigure}[b]{.4\linewidth}
         \centering
         \includegraphics[width=\textwidth]{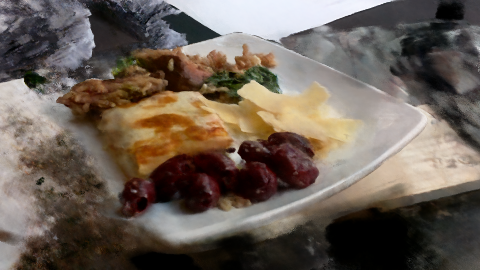}
         \setcounter{subfigure}{6}%
         \caption{Dish(61575996)\\PSNR=17.540, SSIM=0.601, LPIPS=0.227}
         \label{fig:r_dish_1561575996}
     \end{subfigure} &
      \begin{subfigure}[b]{.4\linewidth}
         \centering
         \includegraphics[width=\textwidth]{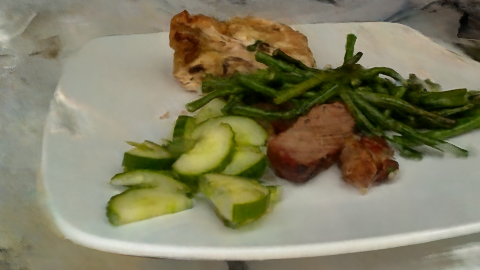}
         \setcounter{subfigure}{7}%
         \caption{Dish(61664061)\\PSNR=22.1, SSIM=0.815, LPIPS=0.119}
         \label{fig:r_dish_1561664061}
     \end{subfigure} \\
\end{tabular}
         \captionsetup{font=small, justification=raggedright, singlelinecheck=false}
        \caption{The qualitative results produced by our framework on the proposed N5k360 dataset. Each scene is accompanied by its associated mean PSNR, SSIM, and LPIPS metrics. Our findings demonstrate that Fig.  ~\ref{fig:r_dish_1561575996} and Fig. ~\ref{fig:r_dish_1561664061} were the most challenging among the eight scenes. This is because their backgrounds contain more information compared to the other scenes, as they feature a paper underneath the dish. Additionally, the N5k360 dataset shares similar challenges with "the wild" \cite{martin2021nerf}.}
        \label{fig:rendered_dishes}
\end{figure*}

 Similarly, Fig \ref{fig:rendered_dishes} highlights the rendered frames of the 8 evaluated dishes from the N5k360 dataset, along with their corresponding evaluation metrics. 
We trained our framework on the N5k360 dataset for 50,000 iterations per scene. Despite the relatively low number of iterations, our model produces high-quality renderings for scenes such as those depicted in Fig. \ref{fig:r_dish_1550704750}, Fig. \ref{fig:r_dish_1550710793}, Fig. \ref{fig:r_dish_1550712459}, Fig. \ref{fig:r_dish_1550772617}, and Fig. \ref{fig:r_dish_1550777256}. Interestingly, for some scenes with a small number of frames, our model shows lower rendering performance, as shown in Fig. \ref{fig:r_dish_1550775219}, Fig. \ref{fig:r_dish_1561575996}, and Fig. \ref{fig:r_dish_1561664061}. In addition, our framework is capable of recognizing foreground and background objects, resulting in well-defined depth maps, as shown in Fig \ref{fig:three graphs}.

Table \ref{table:detailed_results} provides a comprehensive comparison of the Mip-NeRF 360 dataset with state-of-the-art methods, categorized by class. This table also highlights the top three performers for each evaluation metric.
Upon reviewing Table \ref{table:detailed_results}, we observed that our framework, despite having a low number of iterations, is capable of competing with the current state of the art. Additionally, it has also exhibited a higher PSNR in 5 of the 9 scenes (3 outdoor and 2 indoor). Moreover, it is worth noting that our model experiences a degradation in performance when handling the flowers and treehill scenes that contain intricate details in the background, such as grass, tree leaves, and highly textured grounds. Furthermore, our analysis of the SSIM reveals that in 5 out of the 9 scenes (3 outdoor and 2 indoor), our framework outperforms all NeRF-like approaches, while it shows comparable performance to the NeRF-like methods in the kitchen (marked in orange) and bicycle (marked in yellow) scenes. For LPIPS, we outperform all NeRF-like methods in five scenes (1 outdoor and 4 indoor) and compete with the Stable View Synthesis (SVS) \cite{riegler2021stable} and Mip-NeRF 360 \cite{barron2022mip} in the other 4 scenes. Our model utilizes a per-pixel reconstruction loss during the training process, whereas SVS employs a self-supervised approach that directly minimizes a perceptual loss, similar to LPIPS. This fundamental difference in the training methodology could explain the observed difference in the results.

For more detailed information, we have provided rendering and depth videos in the supplemental materials.

\begin{table*}[b!]
\begin{center}
\begin{tabular}{|l|ccccc|cccc|}
\multicolumn{10}{c}{PSNR ↑} \\
 \hline & & \multicolumn{3}{c}{Outdoor} & & & \multicolumn{2}{c}{Indoor} & \\  
 & bicycle & flowers & garden & stump & treehill & room & counter & kitchen & bonsai \\ 
 \hline
NeRF \cite{dengjaxnerf, mildenhall2021nerf}  & 21.76 & 19.40 & 23.11 & 21.73 & 21.28 & 28.56 & 25.67 & 26.31 & 26.81 \\
NeRF w/ DONeRF \cite{neff2021donerf} param. & 21.67 & 19.48 & 23.29 & 23.38 & 21.70 & 28.28 & 25.74 & 25.42 & 27.32 \\
mip-NeRF \cite{barron2021mip} & 21.69 & 19.31 & 23.16 & 23.10 & 21.21 & 28.73 & 25.59 & 26.47 & 27.13 \\
NeRF++ \cite{zhang2020nerf++} & 22.64 & 20.31 & 24.32 & 24.34 & \cellcolor{yellow!25} 22.20 & 28.87 & 26.38 & 27.80 & 29.15 \\
Deep Blending \cite{hedman2018deep} & 21.09 & 18.13 & 23.61 & 24.08 & 20.80 & 27.20 & 26.28 & 25.02 & 27.08 \\
Point-Based Neural Rendering \cite{kopanas2021point} & 21.64 & 19.28 & 22.50 & 23.90 & 20.98 & 26.99 & 25.23 & 24.47 & 28.42 \\
Stable View Synthesis \cite{riegler2021stable}  & 22.79 & 20.15 & \cellcolor{yellow!25} 25.99 & 24.39 & 21.72 & 28.93 & 26.40 & 28.49 & 29.07 \\
mip-NeRF \cite{barron2021mip} w/bigger MLP & 22.90 & 20.79 & 25.85 & 23.64 & 21.71 & \cellcolor{yellow!25} 30.67 & \cellcolor{yellow!25} 28.61 & 29.95 & \cellcolor{orange!25} 31.59 \\
NeRF++ \cite{zhang2020nerf++} w/bigger MLPs & 23.75 & \cellcolor{yellow!25} 21.11 & 25.91 & 25.48 & \cellcolor{orange!25} 22.77 & 30.13 & 27.79 & 29.85 & 30.68 \\
Mip-NeRF 360 \cite{barron2022mip} & \cellcolor{orange!25} 24.37 & \cellcolor{red!25} 21.73 & \cellcolor{orange!25} 26.98 & \cellcolor{orange!25} 26.40 & \cellcolor{red!25} 22.87 & \cellcolor{orange!25} 31.63 & \cellcolor{orange!25} 29.55 & \cellcolor{red!25} 32.23 & \cellcolor{red!25} 33.46 \\
Mip-NeRF 360 \cite{barron2022mip} w/GLO & \cellcolor{yellow!25} 23.95 & \cellcolor{orange!25} 21.60 & 25.09 & \cellcolor{yellow!25} 25.98 & 21.99 & 28.24 & 28.40 & \cellcolor{yellow!25} 30.81 & 30.27 \\
\hline
Our Model & \cellcolor{red!25}26.30	& 20.85 &	\cellcolor{red!25} 27.24	& \cellcolor{red!25} 28.61 &	16.73	& \cellcolor{red!25} 31.84	& \cellcolor{red!25} 29.40 &	\cellcolor{orange!25} 31.086	& \cellcolor{yellow!25} 30.54 \\
\hline
\end{tabular}

\begin{tabular}{|l|ccccc|cccc|}
\multicolumn{10}{c}{SSIM ↑} \\
 \hline & & \multicolumn{3}{c}{Outdoor} & & & \multicolumn{2}{c}{Indoor} & \\  
 & bicycle & flowers & garden & stump & treehill & room & counter & kitchen & bonsai \\ 
 \hline
NeRF \cite{dengjaxnerf, mildenhall2021nerf} & 0.455 & 0.376 & 0.546 & 0.453 & 0.459 & 0.843 & 0.775 & 0.749 & 0.792 \\
NeRF w/ DONeRF \cite{neff2021donerf} param. & 0.454 & 0.379 & 0.542 & 0.522 & 0.461 & 0.841 & 0.776 & 0.678 & 0.813 \\
mip-NeRF \cite{barron2021mip} & 0.454 & 0.373 & 0.543 & 0.517 & 0.466 & 0.851 & 0.779 & 0.745 & 0.818 \\
NeRF++ \cite{zhang2020nerf++} & 0.526 & 0.453 & 0.635 & 0.594 & 0.530 & 0.852 & 0.802 & 0.816 & 0.876 \\
Deep Blending \cite{hedman2018deep} & 0.466 & 0.320 & 0.675 & 0.634 & 0.523 & 0.868 & 0.856 & 0.768 & 0.883 \\
Point-Based Neural Rendering \cite{kopanas2021point} & 0.608 & 0.487 & 0.735 & 0.651 & 0.579 & 0.887 & 0.868 & 0.876 & 0.919 \\
Stable View Synthesis \cite{riegler2021stable} & 0.663 & 0.541 & 0.818 & 0.683 & \cellcolor{yellow!25} 0.606 & 0.905 & 0.886 & 0.910 & 0.925 \\
mip-NeRF \cite{barron2021mip} w/bigger MLP & 0.612 & 0.514 & 0.777 & 0.643 & 0.577 & 0.903 & 0.877 & 0.902 & \cellcolor{yellow!25} 0.928 \\
NeRF++ \cite{zhang2020nerf++} w/bigger MLPs & 0.630 & 0.533 & 0.761 & 0.687 & 0.597 & 0.883 & 0.857 & 0.888 & 0.913 \\
Mip-NeRF 360 \cite{barron2022mip} & \cellcolor{orange!25} 0.685 & \cellcolor{orange!25} 0.583 & \cellcolor{orange!25} 0.813 & \cellcolor{yellow!25} 0.744 & \cellcolor{red!25} 0.632 &\cellcolor{orange!25} 0.913 & \cellcolor{orange!25} 0.894 & \cellcolor{red!25} 0.920 & \cellcolor{red!25} 0.941 \\
Mip-NeRF 360 \cite{barron2022mip} w/GLO & \cellcolor{red!25} 0.687 & \cellcolor{yellow!25} 0.582 & \cellcolor{yellow!25} 0.800 & \cellcolor{orange!25} 0.745 & \cellcolor{orange!25} 0.619 &  \cellcolor{yellow!25} 0.907 & \cellcolor{yellow!25} 0.890 & \cellcolor{yellow!25} 0.916 & \cellcolor{orange!25} 0.932 \\ \hline
Our Model & \cellcolor{yellow!25} 0.670 & \cellcolor{red!25} 0.603 & \cellcolor{red!25} 0.824 & \cellcolor{red!25} 0.810 & 0.420 & \cellcolor{red!25} 0.931 & \cellcolor{red!25} 0.904 & \cellcolor{orange!25} 0.918 & 0.910 \\
\hline
\end{tabular}

\begin{tabular}{|l|ccccc|cccc|}
\multicolumn{10}{c}{LPIPS ↓} \\
 \hline & & \multicolumn{3}{c}{Outdoor} & & & \multicolumn{2}{c}{Indoor} & \\  
 & bicycle & flowers & garden & stump & treehill & room & counter & kitchen & bonsai \\ 
 \hline
NeRF \cite{dengjaxnerf, mildenhall2021nerf} & 0.536 & 0.529 & 0.415 & 0.551 & 0.546 & 0.353 & 0.394 & 0.335 & 0.398 \\
NeRF w/ DONeRF \cite{neff2021donerf} param. & 0.542 & 0.539 & 0.436 & 0.492 & 0.545 & 0.368 & 0.394 & 0.410 & 0.368 \\
mip-NeRF  \cite{barron2021mip} & 0.541 & 0.535 & 0.422 & 0.490 & 0.538 & 0.346 & 0.390 & 0.336 & 0.370 \\
NeRF++ \cite{zhang2020nerf++} & 0.455 & 0.466 & 0.331 & 0.416 & 0.466 & 0.335 & 0.351 & 0.260 & 0.291 \\
Deep Blending \cite{hedman2018deep} & 0.377 & 0.476 & 0.231 & 0.351 & 0.383 & 0.266 & 0.258 & 0.246 & 0.275 \\
Point-Based Neural Rendering \cite{kopanas2021point} & 0.313 & 0.372 & 0.197 & 0.303 & \cellcolor{orange!25} 0.325 & 0.216 & 0.209 & 0.160 & \cellcolor{yellow!25} 0.178 \\
Stable View Synthesis \cite{riegler2021stable} & \cellcolor{red!25} 0.243 & \cellcolor{red!25} 0.317 & \cellcolor{red!25} 0.137 & 0.281 & \cellcolor{red!25} 0.286 &  \cellcolor{orange!25} 0.182 & \cellcolor{orange!25} 0.168 & \cellcolor{orange!25} 0.125 & \cellcolor{orange!25} 0.164 \\
mip-NeRF \cite{barron2021mip} w/bigger MLP & 0.372 & 0.407 & 0.205 & 0.357 & 0.401 & 0.229 & 0.239 & 0.152 & 0.204 \\
NeRF++ \cite{zhang2020nerf++} w/bigger MLPs & 0.356 & 0.395 & 0.223 & 0.328 & 0.386 & 0.270 & 0.270 & 0.177 & 0.230 \\
Mip-NeRF 360 \cite{barron2022mip} & \cellcolor{yellow!25} 0.301 &\cellcolor{yellow!25} 0.344 & \cellcolor{orange!25} 0.170 & \cellcolor{yellow!25} 0.261 & 0.339 & \cellcolor{yellow!25} 0.211 & \cellcolor{yellow!25} 0.204 & \cellcolor{yellow!25} 0.127 & 0.176 \\
Mip-NeRF 360 \cite{barron2022mip} w/GLO & \cellcolor{orange!25} 0.296 & \cellcolor{orange!25} 0.343 & \cellcolor{yellow!25} 0.173 & \cellcolor{orange!25} 0.258 & \cellcolor{yellow!25} 0.338 & 0.208 & 0.206 & 0.129 & 0.182 \\ 
\hline
Our Model & 0.334 & 0.350 & 0.174 & \cellcolor{red!25} 0.214 & 0.511 &  \cellcolor{red!25} 0.047 &  \cellcolor{red!25} 0.078 &  \cellcolor{red!25} 0.062 &  \cellcolor{red!25} 0.036 \\
\hline
\end{tabular}
\end{center}
\caption{A detailed version of Table ~\ref{table:avg_comparasion} in the main paper, which provides metrics for individual scenes and assesses the performance of our framework alongside various NeRF and non-NeRF baselines. Even though some scenes are more challenging than others, the relative rankings of all techniques across each scene tend to align with the average metrics ranking. The top result is denoted by the color red, the second by orange, and the third by yellow.}
\label{table:detailed_results}
\end{table*}

\end{appendices}

\end{document}